\documentclass[10pt,twocolumn,letterpaper]{article}

\usepackage[pagenumbers]{cvpr} % To force page numbers, e.g. for an arXiv version

\usepackage{amsmath,amssymb,booktabs,multirow,graphicx,array}
\usepackage{colortbl}
\usepackage{arydshln}
\usepackage{siunitx}
\usepackage{fontawesome5}
\definecolor{tabfirst}{rgb}{1, 0.7, 0.7}
\definecolor{tabsecond}{rgb}{1, 0.85, 0.7}
\definecolor{tabthird}{rgb}{1, 1, 0.7}
\definecolor{gr}{rgb}{0.75,0.95,0.85}
\definecolor{best}{rgb}{0.4, 0.4, 0.8}
\usepackage{xcolor}
\usepackage{colortbl}
\definecolor{DualSpecBlue}{RGB}{235,244,252}

\usepackage{dashrule}

\newcommand{\ours}{DualSpec}

\definecolor{cvprblue}{rgb}{0.21,0.49,0.74}
\usepackage[pagebackref,breaklinks,colorlinks,allcolors=cvprblue]{hyperref}

\def\confName{CVPR}
\def\confYear{2027}

\title{DualSpec: Spectral Reconstruction with Optimized Stereo Cameras}

\author{
Luca Cogo\quad
Marco Buzzelli\quad
Simone Bianco\quad
Raimondo Schettini\\
University of Milano-Bicocca\\
{\tt\small \{\href{mailto:luca.cogo@unimib.it}{\textcolor{black}{luca.cogo}}, \href{mailto:marco.buzzelli@unimib.it}{\textcolor{black}{marco.buzzelli}}, \href{mailto:simone.bianco@unimib.it}{\textcolor{black}{simone.bianco}}, \href{mailto:raimondo.schettini@unimib.it}{\textcolor{black}{raimondo.schettini}}\}@unimib.it} 
\\[0.5em]
{\small \href{https://lucacogo.github.io/DualSpec/}{\faGlobe\enspace Project Page}}}

\begin{document}
\maketitle
\begin{abstract}
Spectral imaging captures information that ordinary RGB sensors cannot, but dedicated spectral cameras can be expensive and bulky. Recovering dense spectra from RGB images offers an accessible alternative, but distinct spectra can produce the same camera response, generating an ambiguity that even more advanced recovery algorithms cannot always resolve. Therefore, increasing the spectral information captured at acquisition is essential.
We propose \ours{}, a stereo acquisition framework that jointly optimizes two external optical filters and a reconstruction network. Each filter is placed in front of a standard RGB camera, and the two are designed to produce six responses that provide complementary information for the network.
Experiments across reconstruction architectures and spectral ranges show that \ours{} reduces mean spectral angle error by 36--72\% relative to a single RGB camera and by 11--29\% relative to stereo pairs selected from commercial filters.
Cross-dataset evaluations, including a new benchmark of spectral images captured from displaced viewpoints, show that the method generalizes and remains robust to viewpoint misalignment, while analyses of thin-film approximations support optical feasibility.

\end{abstract}

\begin{figure}[t]
\centering
\includegraphics[width=\linewidth]{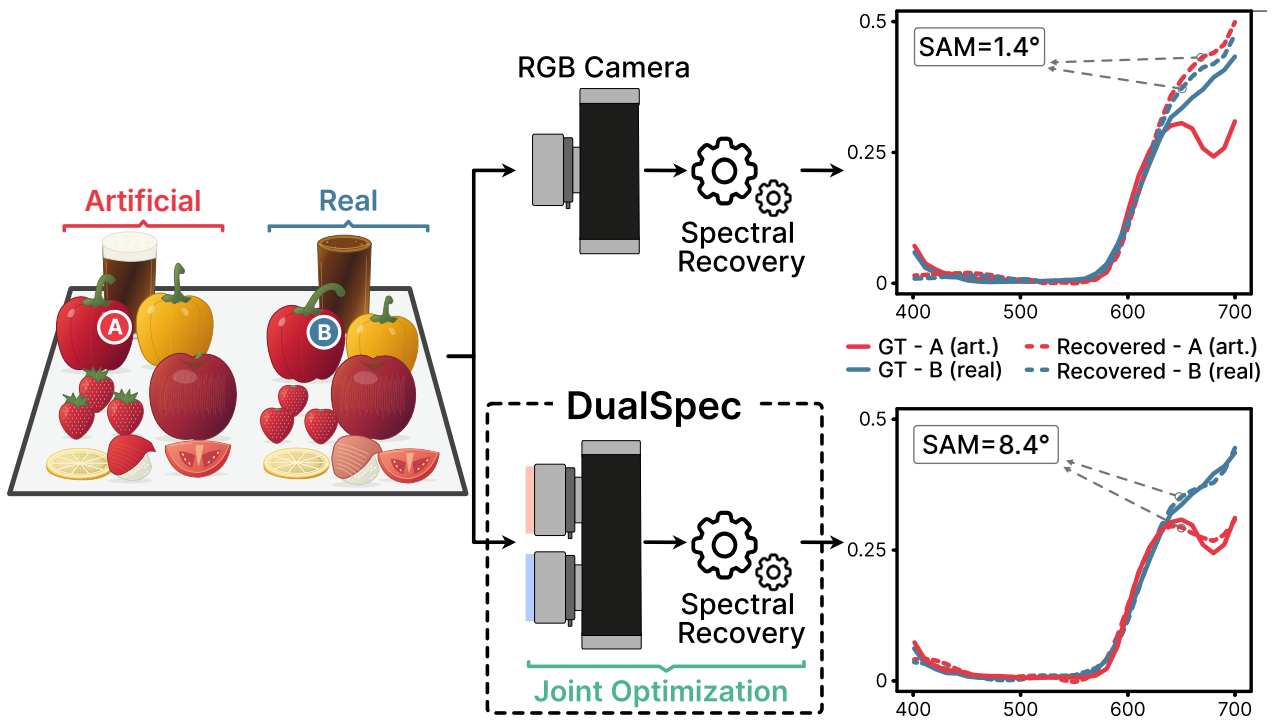}
\caption{Cross-dataset reconstruction with MST++~\cite{cai2022mst++} on a scene from the CAVE dataset~\cite{yasuma2010generalized}. The artificial (A) and real (B) peppers look similar in RGB but have distinct spectra. 
The complementary measurements acquired by \ours{} yield reconstructions that better distinguish the spectral signatures of the two samples than those obtained with a single RGB camera. Solid and dashed curves represent ground-truth and reconstructed spectra, respectively. The reported spectral angle (SAM) measures the separation between the two reconstructed spectra.}
\label{fig:overview}
\end{figure}

\section{Introduction}\label{sec:intro}

Spectral imaging captures how light from a scene varies across wavelengths, preserving details that get lost when measurements are reduced to three RGB channels. This additional information supports applications in remote sensing, healthcare, industrial inspection, and computational photography~\cite{hong2026hyperspectral,hong2024spectralgpt,leon2023hyperspectral,yang2025hyperspectral,cogo2026leveraging}. Hyperspectral cameras measure spectra directly, but often rely on scanning mechanisms that acquire scenes sequentially, limiting their use in dynamic settings and requiring bulky optical and mechanical components. Snapshot systems avoid sequential capture by using filter arrays or coded optics, but can trade spatial resolution for spectral resolution or introduce more elaborate optics~\cite{miao2006design,wagadarikar2008single}. Overall, the cost and complexity of these specialized imaging systems remain significant barriers to their large-scale adoption. These limitations motivate spectral reconstruction from RGB images as a more accessible alternative for obtaining dense spectral information with widely available cameras.

Recovering a spectrum from an RGB measurement is, however, fundamentally ill-posed because each RGB channel integrates the incident spectrum through a broad sensitivity function. In particular, metamerism occurs when distinct spectra produce identical RGB responses under the same imaging conditions. A learned reconstruction model can use statistical priors in natural scenes to predict a likely spectrum, but the RGB values alone cannot identify which of two metamers generated the measurement~\cite{fu2025limitations}. 
\Cref{fig:overview} illustrates this ambiguity using real and artificial objects whose similar RGB appearances can make their distinct spectra difficult to tell apart in a single-camera reconstruction.

Reducing this ambiguity requires additional, spectrally diverse measurements. Adding an external filter attachment for a single camera~\cite{nie2018deeply,zhang2024learning} can reshape its sensitivities, but three measurements per pixel remain subject to metameric ambiguity. Multi-camera systems instead capture the scene through several standard RGB cameras, each behind a different external optical filter. The filters give the views complementary spectral sensitivities, so that two cameras yield six measurements at each pixel and constrain the underlying spectrum more strongly than a single RGB triplet. Previous multi-camera approaches select these filters from commercial catalogs~\cite{shrestha2011multispectral,reutsky2025modulate}. Such filters are readily available, but the catalog restricts the transmittances that can be combined. Moreover, because training a reconstruction network for every candidate pair is impractical, selection relies on proxy criteria, and the chosen pair may be suboptimal for the network that ultimately processes its measurements.

Rather than selecting filters from a fixed catalog, we design them together with the network that uses their measurements. We propose \ours{}, a stereo spectral reconstruction framework that jointly optimizes the transmittance of an external attachment for each RGB camera and the weights of the reconstruction network. A differentiable image-formation model propagates the reconstruction loss through both filtered views, allowing the attachments to develop complementary responses tailored to the network. The model accounts for light loss through each filter, so that optimization balances spectral selectivity against noise. To enforce physical manufacturability, we parameterize each transmittance with a Gaussian basis and a sigmoid, yielding smooth curves with values between zero and one.

Our contributions can be summarized as follows:
\begin{itemize}
\item We introduce \ours{}, a stereo spectral acquisition framework that jointly optimizes two attachments and a spectral reconstruction network for standard RGB cameras.
\item We show consistent gains across backbones and spectral ranges over single-camera and commercial-filter stereo baselines.
\item We collect a displaced-view benchmark of spectral image pairs captured from different viewpoints, on which the learned designs transfer without fine-tuning.
\item We assess optical feasibility through multilayer thin-film approximations, showing that a short network fine-tuning on the fitted transmittances recovers the performance of the learned designs.
\end{itemize}
\section{Related Work}\label{sec:related}

\paragraph{Spectral image acquisition.}
Several acquisition designs have been proposed to capture dense spectral information, each introducing different trade-offs in acquisition speed, spatial resolution, hardware complexity, or operating conditions. Line-scan instruments and filter-wheel cameras provide finely sampled spectral measurements through sequential acquisition, but this increases capture time, sensitivity to scene motion, and optical--mechanical complexity~\cite{vane1984airborne,brauers2008multispectral,behmann2018specim}. 
Snapshot multispectral filter-array systems capture all bands in a single exposure, but allocate different wavelengths to different pixels, producing a trade-off between spatial and spectral resolution~\cite{miao2006design,thomas2025trends}. 

Controlled-illumination methods efficiently acquire complementary spectral measurements through multiplexed lighting, rolling-shutter acquisition, or coded exposures~\cite{park2007multispectral,verma2026lumosaic}. However, their reliance on synchronized active lighting largely restricts their use to controlled indoor settings.

These approaches demonstrate different ways of acquiring informative spectral cues. However, their reliance on specialized sensors, optics, scanning mechanisms, or active illumination makes spectral imaging systems costly and limits their widespread adoption. We therefore target a more accessible stereo device that uses standard RGB cameras equipped with passive external attachments of different transmittances, providing a path toward low-cost snapshot spectral acquisition.

\vspace{-2mm}
\paragraph{Spectral recovery from RGB.}
Early RGB-to-spectrum methods operated independently on individual pixels, constraining the set of possible spectra using low-dimensional bases, regression, or sparse representations~\cite{parkkinen1989characteristic,nguyen2014training,arad2016sparse}. As larger datasets and public spectral-recovery challenges became available, research shifted toward deep convolutional neural networks that exploit spatial context~\cite{lim2017enhanced,shi2018hscnn,arad2020ntire,arad2022ntire}. Subsequent approaches incorporated attention layers, including spectral-wise attention and spatial--spectral attention mechanisms, to capture dependencies across spectral channels and spatial locations~\cite{cai2022mst++,yang2024hyperspectral,yao2024specat}. More recent work explores generative spectral priors to guide the recovery of spectral information from RGB observations~\cite{yang2026unraveling}. Although this progression has substantially improved reconstruction accuracy, increasing model complexity or strengthening the learned prior does not increase the spectral information contained in the RGB input. Because distinct spectra can produce the same camera response, a reconstruction model can only use priors learned from the training distribution to select a likely solution and cannot uniquely distinguish metamers from the RGB measurements alone~\cite{fu2025limitations,yang2026unraveling}. 

Rather than introducing another reconstruction architecture, our approach increases the information available at acquisition through complementary optical measurements, revealing spectral differences that a single RGB observation might miss.

\vspace{-2mm}
\paragraph{Optimizing spectral responses.}
Optical filtering modifies a camera's effective spectral sensitivity without replacing its sensor. In camera color correction, chromagenic approaches optimize an external filter to improve the mapping from RGB measurements to device-independent color coordinates~\cite{finlayson2021designing}. A similar strategy has been applied to spectral reconstruction by selecting an attachment from a bank of measured filters~\cite{arad2017filter}. A suitable filter can make spectra that appear similar under the native camera sensitivities more distinguishable. However, a predetermined filter bank restricts the design space and makes reconstruction performance dependent on the available candidates. Recent methods expand this space by jointly adapting acquisition and reconstruction through selected or learned camera sensitivities, or optical attachments~\cite{fu2018joint,nie2018deeply,sun2021tuning,zhang2024learning}. While these methods demonstrate the benefit of tailoring spectral responses to the reconstruction task, they either retain only three measurements from a single camera or require modifications to the sensor hardware.

Multi-camera approaches acquire additional spectral information by placing different filters in front of multiple RGB sensors. Shrestha \etal~\cite{shrestha2011multispectral} select physically available filter pairs through exhaustive evaluation using shallow regression models. 
Reutsky~\etal~\cite{reutsky2025modulate} select filters for a multi-camera smartphone by minimizing spectral uncertainty and then learn reconstruction from the resulting misaligned views. In both cases, the filters are fixed before the network is trained and are drawn from a candidate bank. Exhaustive pairwise search also scales quadratically with the number of filters, which forces reliance on proxy criteria that may not align with final reconstruction performance. \ours{} instead learns the transmittances jointly with the network, allowing each to adapt to the other.

\section{DualSpec}\label{sec:method}

\ours{} increases the information available for spectral reconstruction by capturing a scene with two RGB cameras equipped with different passive optical attachments. We jointly learn the attachment transmittances and the reconstruction network so that the two cameras acquire complementary measurements tailored to spectral recovery.

\subsection{Two filtered RGB views}\label{sec:camera}

Our acquisition model uses two cameras with the same unmodified RGB sensor model, so their measurements differ only because of their external attachments. Let $(x,y)$ denote a pixel location, $\lambda$ a wavelength, $r(x,y,\lambda)$ the scene reflectance, $e(\lambda)$ the illuminant spectrum, and $s_c(\lambda)$ the native sensitivity of RGB channel $c\in\{r,g,b\}$. For camera $j\in\{1,2\}$, an attachment with transmittance $a_j(\lambda)$ produces the response
\begin{equation}\label{eq:forward}
i_{j,c}(x,y)
=
\int_{\lambda_{\min}}^{\lambda_{\max}}
e(\lambda)r(x,y,\lambda)s_c(\lambda)a_j(\lambda)
\,\mathrm d\lambda,
\end{equation}
where $i_{j,c}(x,y)$ is the intensity recorded at pixel $(x,y)$ by channel $c$ of camera $j$, and $[\lambda_{\min},\lambda_{\max}]$ is the spectral acquisition range. The recorded intensity is the scene spectrum integrated over this range and weighted by the sensor sensitivity and attachment transmittance.

The attachments modify the effective channel sensitivities from $s_c$ to $s_ca_1$ and $s_ca_2$. Let $\mathbf I_j\in\mathbb R^{H\times W\times 3}$ denote the RGB image captured by camera $j$, where $H$ and $W$ are the image height and width. We concatenate the two images along their channel dimension to obtain $\mathbf I\in\mathbb R^{H\times W\times 6}$. A reconstruction network $f_{\boldsymbol{\theta}}$, parameterized by the vector $\boldsymbol{\theta}$, maps this input to the reconstructed spectral image $\widehat{\mathbf R}=f_{\boldsymbol{\theta}}(\mathbf I)\in\mathbb R^{H\times W\times K}$, where $K$ is the number of reconstructed wavelength bands. Single-camera configurations instead provide one three-channel image $\mathbf I_j$ to the network.

We select the sensor sensitivities according to the spectral range of each experiment. Visible-range experiments use measured sensitivities for a Sony IMX477 sensor~\cite{sony_imx477} with its IR-cut filter. Because the available IMX477 measurements do not extend into the near-infrared, experiments covering 400--1000\,nm use a sensitivity model for a Sony IMX224 sensor~\cite{sony_imx224} without an IR-cut filter, which spans the full spectral interval.

\paragraph{Light efficiency.}
We use capped auto-exposure to account for light lost through each attachment. We set the exposure of a reference unfiltered camera to collect a fixed mean number of photoelectrons in its green channel. Each filtered camera increases its exposure to match that mean, up to a factor of $t_{\max}$. If the exposure cap is reached before the target is met, the camera collects fewer photoelectrons and has a lower signal-to-noise ratio. We then add Poisson shot noise and Gaussian dark noise and quantize the resulting signal. Finally, we divide each channel's values by its response to an equal-energy white spectrum to simulate digital gain.

\subsection{Joint optical--computational learning}\label{sec:objective}

We jointly optimize the attachment transmittances and the network parameters $\boldsymbol{\theta}$. Let $\mathbf R\in\mathbb R^{H\times W\times K}$ denote the target spectral image corresponding to the reconstruction $\widehat{\mathbf R}$. We define the reconstruction loss as
\begin{equation}\label{eq:mae}
\mathcal L_{\mathrm{rec}}
=
\operatorname{mean}
\left(
\left|\mathbf R-\widehat{\mathbf R}\right|
\right).
\end{equation}
The absolute value is applied element-wise, and $\operatorname{mean}(\cdot)$ averages all elements of its argument. Therefore, $\mathcal L_{\mathrm{rec}}$ is the mean absolute error over all spatial locations and wavelength bands.

\paragraph{Attachment parameterization.}
To enforce physical manufacturability of the filters, we parameterize each attachment’s spectral transmittance using a smooth basis. Let $g_1,\dots,g_B$ be Gaussian basis functions with standard deviation $\sigma_g$ and centers spaced by $\Delta_g$ across the spectral range. The transmittance of attachment $j$ is
\begin{equation}\label{eq:basis}
a_j(\lambda)
=
\operatorname{sig}\!\left(
\gamma\left(\sum_{b=1}^{B} w_{j,b}\,g_b(\lambda)+c_j\right)
\right),
\end{equation}
where $\operatorname{sig}$ is the logistic function, $w_{j,b}$ and $c_j$ are learned coefficients, and $\gamma$ is a fixed gain. The logistic function constrains transmittance to $(0,1)$, while the Gaussian basis keeps the curves smooth and controls the scale of their spectral features.

We use the learned curves as absolute transmittances, so the amount of light each filter passes determines the exposure and noise in its view under the acquisition model in~\Cref{sec:camera}. 

\begin{figure}[t]
\centering
\includegraphics[width=\columnwidth]{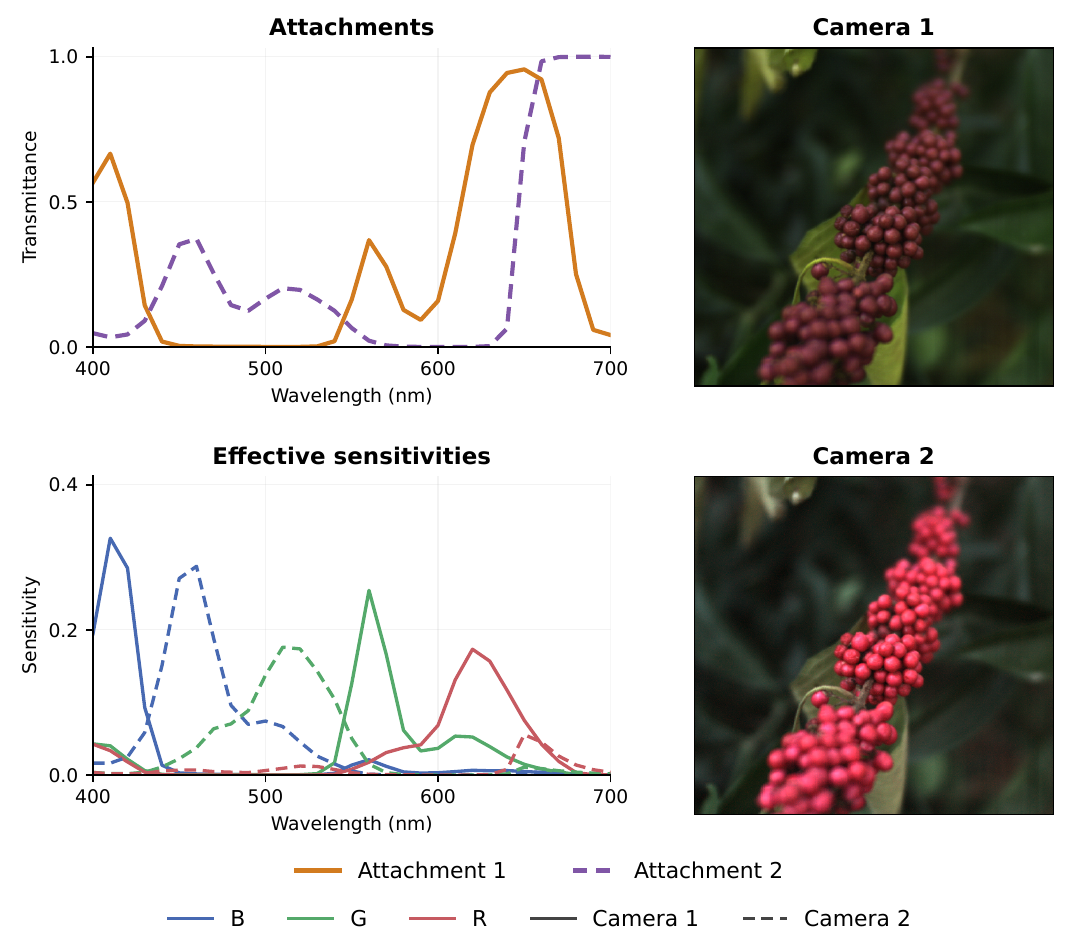}
\caption{Attachments learned with MST++~\cite{cai2022mst++} (top left), effective sensitivities (bottom left), and the resulting network inputs for an Arad-1K scene. Solid and dashed curves correspond to the first and second cameras. 
}
\label{fig:attachments_mstpp}
\end{figure}

As shown in~\Cref{fig:attachments_mstpp}, the attachments develop distinct spectral responses rather than reproducing the same measurements, so the two views of a scene differ in color. The attachments learned with the other backbones are shown in the supplementary material.

\subsection{Two-stage optimization}\label{sec:training}

Optimization proceeds in two stages. In the first stage, we jointly optimize the attachment coefficients and the network parameters $\boldsymbol{\theta}$ through the acquisition model described in~\Cref{sec:camera}. The shot noise is replaced by its Gaussian approximation, with variance equal to the collected charge, so that gradients of~\cref{eq:mae} propagate through noisy measurements and the attachments are optimized for the signal-to-noise ratio they actually produce. 

In the second stage, we freeze the selected attachments and train the reconstruction network from scratch with the exact acquisition model, including Poisson shot noise, Gaussian dark noise, and quantization~\cite{arad2020ntire}. Separating the two stages avoids differentiating through Poisson sampling and lets the final network adapt to the noise statistics of the fixed optical design.

The framework applies to any differentiable reconstruction network. We evaluate three representative families: a pixel-wise MLP, the convolutional EDSR~\cite{lim2017enhanced}, and the attention-based MST++~\cite{cai2022mst++}, representing per-pixel spectral mapping, spatial convolution, and spatial--spectral attention. Since only the input layer depends on the number of channels, \ours{} adds negligible inference cost over a single camera.
\section{Experiments}\label{sec:experiments}

\subsection{Experimental setup}\label{sec:setup}

\paragraph{Datasets.}\label{sec:data}
We use the Arad-1K dataset~\cite{arad2022ntire} as our primary visible-range benchmark. It features 950 publicly available natural scenes with a spatial resolution of $482\times512$ pixels and 31 bands spanning 400--700\,nm. We divide them into training, validation, and test sets using an 80:10:10 split. For VNIR reconstruction, we use the 201 scenes of the BGU-ICVL dataset~\cite{arad2016sparse}. We resample cubes to 61 bands spanning 400--1000\,nm and resize them to a width of 512 pixels. We reserve 20\% of these scenes for testing and use 20\% of the remaining scenes for validation.

To evaluate cross-dataset generalization and robustness to viewpoint misalignment, we introduce a displaced-view benchmark containing ten pairs of spectral images independently captured from displaced viewpoints using a Specim IQ camera~\cite{specimiq}. The benchmark comprises five staged indoor scenes and five outdoor scenes, spanning controlled compositions and natural environments. Further details about the acquisition setup and dataset composition are provided in the supplementary material.

\paragraph{Baselines.}\label{sec}
We compare \ours{} with an unfiltered single RGB camera and a single-camera configuration with one learned attachment, following the joint optimization approach of Nie \etal~\cite{nie2018deeply}. For a fair comparison, the learned single attachment uses the same parameterization and acquisition model as \ours{}. For the stereo setting, we compare with adaptations of the filter-selection criteria proposed by Shrestha \etal~\cite{shrestha2011multispectral} and Reutsky \etal~\cite{reutsky2025modulate}. To this end, we construct two candidate filter banks from manufacturer-published transmittance curves, comprising 210 responses collected from the Lee Filters catalog~\cite{lee_filters_catalog} and 136 from the Omega Optical catalog~\cite{omega_filters_catalog}. When forming candidate pairs, we also allow one camera to remain unfiltered, resulting in 22,155 candidate pairs for Lee and 9,316 for Omega Optical. Both selections use the same sensor as \ours{}. Once a pair is selected, its responses are fixed and the corresponding MLP, EDSR, or MST++ reconstructor is trained using the same protocol as the second stage of \ours{}. Additional details about catalog preparation and filter selection baselines are provided in the supplementary material.\vspace{-2mm}

\begin{table}[t]
\caption{Spectral reconstruction on Arad-1K. Superscripts identify the filter bank used by each selection method. Bold values indicate the best result within each backbone and statistic, while blue shading identifies \ours.\vspace{-2mm}}\label{tab:arad}
\centering\scriptsize
\setlength{\tabcolsep}{1.8pt}
\resizebox{\columnwidth}{!}{%
\begin{tabular}{ll*{9}{c}}
\toprule
\multirow[c]{2}{*}[-2.5pt]{Backbone} &
\multirow[c]{2}{*}[-2.5pt]{Acquisition / selection} &
\multicolumn{3}{c}{SAM ($^\circ$) $\downarrow$} &
\multicolumn{3}{c}{RMSE $\times 1000\,\downarrow$} &
\multicolumn{3}{c}{PSNR $\uparrow$} \\
\cmidrule(lr){3-5}\cmidrule(lr){6-8}\cmidrule(l){9-11}
 & & Mean & Median & P95 & Mean & Median & P95 & Mean & Median & P05 \\
\midrule
\multirow[c]{7}{*}[-3.5pt]{MLP}
 & Single RGB & 7.89 & 6.04 & 21.05 & 25.4 & 16.7 & 75.0 & 35.26 & 35.54 & 22.50 \\
 & Single + learned filter & 6.21 & 4.81 & 16.61 & 16.7 & 11.0 & 51.1 & 38.62 & 39.18 & 25.82 \\
\noalign{\vskip 2pt}
\cdashline{2-11}
\noalign{\vskip 2pt}
 & Reutsky$^{\text{Lee}}$ & 5.57 & 4.47 & 13.84 & 14.9 & 11.0 & 38.6 & 39.05 & 39.19 & 28.27 \\
 & Shrestha$^{\text{Lee}}$ & 5.81 & 4.52 & 16.07 & 17.3 & 11.6 & 48.2 & 38.16 & 38.73 & 26.35 \\
 & Reutsky$^{\text{Omega}}$ & 5.33 & 4.39 & 12.69 & 13.8 & 10.7 & 36.7 & 39.60 & 39.39 & 28.70 \\
 & Shrestha$^{\text{Omega}}$ & 6.35 & 5.03 & 16.31 & 19.6 & 12.6 & 62.8 & 37.23 & 38.00 & 24.04 \\
\rowcolor{DualSpecBlue}
\cellcolor{white}
 & DualSpec (Ours) & \textbf{4.73} & \textbf{3.99} & \textbf{11.10} & \textbf{10.9} & \textbf{8.4} & \textbf{30.9} & \textbf{41.59} & \textbf{41.50} & \textbf{30.20} \\
\midrule
\multirow[c]{7}{*}[-3.5pt]{EDSR}
 & Single RGB & 5.28 & 3.28 & 18.98 & 19.2 & 9.7 & 69.1 & 38.55 & 40.26 & 23.24 \\
 & Single + learned filter & 4.38 & 3.07 & 14.96 & 13.0 & 7.5 & 47.5 & 41.20 & 42.48 & 26.47 \\
\noalign{\vskip 2pt}
\cdashline{2-11}
\noalign{\vskip 2pt}
 & Reutsky$^{\text{Lee}}$ & 2.39 & 2.00 & 4.86 & 6.4 & 4.6 & 16.4 & 46.36 & 46.82 & 35.70 \\
 & Shrestha$^{\text{Lee}}$ & 2.56 & 2.01 & 5.09 & 6.7 & 5.2 & 17.1 & 45.75 & 45.72 & 35.35 \\
 & Reutsky$^{\text{Omega}}$ & 2.43 & 1.85 & 5.08 & 6.7 & 4.9 & 16.6 & 46.20 & 46.16 & 35.60 \\
 & Shrestha$^{\text{Omega}}$ & 3.06 & 2.40 & 6.65 & 8.3 & 5.9 & 20.5 & 43.94 & 44.60 & 33.75 \\
\rowcolor{DualSpecBlue}
\cellcolor{white}
 & DualSpec (Ours) & \textbf{2.02} & \textbf{1.62} & \textbf{4.28} & \textbf{5.3} & \textbf{4.2} & \textbf{15.7} & \textbf{47.98} & \textbf{47.63} & \textbf{36.07} \\
\midrule
\multirow[c]{7}{*}[-3.5pt]{MST++}
 & Single RGB & 5.35 & 3.33 & 18.11 & 18.5 & 9.6 & 75.9 & 39.14 & 40.34 & 22.40 \\
 & Single + learned filter & 3.69 & 2.90 & 9.10 & 11.0 & 6.9 & 42.6 & 42.17 & 43.19 & 27.41 \\
\noalign{\vskip 2pt}
\cdashline{2-11}
\noalign{\vskip 2pt}
 & Reutsky$^{\text{Lee}}$ & 2.28 & 1.98 & 4.46 & 5.9 & 4.6 & 13.1 & 46.51 & 46.80 & 37.74 \\
 & Shrestha$^{\text{Lee}}$ & 2.38 & 2.02 & 4.37 & 6.3 & 4.8 & 17.2 & 46.00 & 46.31 & 35.27 \\
 & Reutsky$^{\text{Omega}}$ & 2.41 & 1.85 & 4.37 & 6.5 & 4.7 & 16.3 & 46.28 & 46.60 & 35.75 \\
 & Shrestha$^{\text{Omega}}$ & 2.77 & 2.35 & 5.38 & 7.6 & 5.7 & 18.9 & 44.55 & 44.81 & 34.47 \\
\rowcolor{DualSpecBlue}
\cellcolor{white}
 & DualSpec (Ours) & \textbf{1.93} & \textbf{1.61} & \textbf{3.59} & \textbf{5.0} & \textbf{3.9} & \textbf{12.9} & \textbf{48.17} & \textbf{48.10} & \textbf{37.78} \\
\bottomrule
\end{tabular}}
\end{table}

\paragraph{Implementation and evaluation.}
The pixel-wise MLP has three hidden layers of widths $(128,512,128)$. We use the original EDSR~\cite{lim2017enhanced} and MST++~\cite{cai2022mst++} architectures, adapting only the input and output layers of each backbone to the number of input channels and output bands. 
All learned attachments use Gaussian bases with $\Delta_g=\sigma_g=10$\,nm and $B=31$ for the visible range (400--700\,nm) or $B=61$ for the VNIR range (400--1000\,nm). We set the transmittance gain to $\gamma=4$ and use a shared learning rate for the attachment coefficients and network weights during joint training.

For the acquisition model described in~\Cref{sec:camera}, we set the unfiltered green-channel target to 5000 photoelectrons, the exposure cap to $t_{\max}=8$, and the dark-noise standard deviation to 100 photoelectrons, with 16-bit quantization~\cite{arad2020ntire}. We train with Adam~\cite{kingma2015adam} for 100 and 300 epochs in the two stages, with initial learning rates of $2\times10^{-3}$ and $4\times10^{-4}$, respectively. Both stages use cosine annealing~\cite{loshchilov2017sgdr} and random $128\times128$ crops. All the experiments in all configurations share the same test-time Poisson noise realization. We evaluate each image using SAM (in degrees), RMSE, and PSNR. For each metric, we summarize the results across images using the mean, median, and 95th percentile (5th for PSNR).

\subsection{Results}
\begin{table}[t]
\caption{Spectral reconstruction on BGU-ICVL over 400--1000\,nm. Commercial-filter baselines are omitted because their measured responses do not extend beyond 700\,nm. Bold values indicate the best result within each backbone and statistic, while blue shading identifies \ours.\vspace{-2mm}}\label{tab:vnir}
\centering\scriptsize
\setlength{\tabcolsep}{1.8pt}
\resizebox{\columnwidth}{!}{%
\begin{tabular}{ll*{9}{c}}
\toprule
\multirow[c]{2}{*}[-2.5pt]{Backbone} &
\multirow[c]{2}{*}[-2.5pt]{Acquisition / selection} &
\multicolumn{3}{c}{SAM ($^\circ$) $\downarrow$} &
\multicolumn{3}{c}{RMSE $\times 1000\,\downarrow$} &
\multicolumn{3}{c}{PSNR $\uparrow$} \\
\cmidrule(lr){3-5}\cmidrule(lr){6-8}\cmidrule(l){9-11}
 & & Mean & Median & P95 & Mean & Median & P95 & Mean & Median & P05 \\
\midrule
\multirow[c]{3}{*}[-3.5pt]{MLP}
 & Single RGB & 8.53 & 6.61 & 20.70 & 26.0 & 18.3 & 93.8 & 30.94 & 32.85 & 20.57 \\
 & Single + learned filter & 3.31 & 3.13 & 6.87 & 10.8 & 9.6 & 22.5 & 36.99 & 37.08 & 32.96 \\
\noalign{\vskip 2pt}
\cdashline{2-11}
\noalign{\vskip 2pt}
\rowcolor{DualSpecBlue}
\cellcolor{white}
 & DualSpec (Ours) & \textbf{2.35} & \textbf{2.15} & \textbf{4.81} & \textbf{6.9} & \textbf{6.3} & \textbf{15.3} & \textbf{40.80} & \textbf{40.80} & \textbf{36.32} \\
\midrule
\multirow[c]{3}{*}[-3.5pt]{EDSR}
 & Single RGB & 2.75 & 2.82 & 4.66 & 9.2 & 8.8 & 16.9 & 38.82 & 38.98 & 33.03 \\
 & Single + learned filter & 2.33 & 2.22 & 3.60 & 7.9 & 7.5 & 16.0 & 39.69 & 39.81 & 34.86 \\
\noalign{\vskip 2pt}
\cdashline{2-11}
\noalign{\vskip 2pt}
\rowcolor{DualSpecBlue}
\cellcolor{white}
 & DualSpec (Ours) & \textbf{1.75} & \textbf{1.80} & \textbf{2.56} & \textbf{5.2} & \textbf{4.9} & \textbf{9.5} & \textbf{43.36} & \textbf{43.65} & \textbf{39.50} \\
\midrule
\multirow[c]{3}{*}[-3.5pt]{MST++}
 & Single RGB & 2.72 & 2.73 & 4.39 & 9.1 & 9.0 & 18.4 & 38.91 & 38.63 & 32.30 \\
 & Single + learned filter & 2.23 & 2.17 & 3.56 & 7.6 & 7.7 & 14.2 & 39.93 & 40.35 & 35.15 \\
\noalign{\vskip 2pt}
\cdashline{2-11}
\noalign{\vskip 2pt}
\rowcolor{DualSpecBlue}
\cellcolor{white}
 & DualSpec (Ours) & \textbf{1.74} & \textbf{1.73} & \textbf{2.65} & \textbf{5.3} & \textbf{5.3} & \textbf{10.5} & \textbf{43.13} & \textbf{43.26} & \textbf{39.34} \\
\bottomrule
\end{tabular}}\vspace{-2mm}
\end{table}

\paragraph{Visible-range reconstruction on Arad-1K.}\label{sec:arad}
\begin{figure*}[t]
\centering
\includegraphics[width=\textwidth]{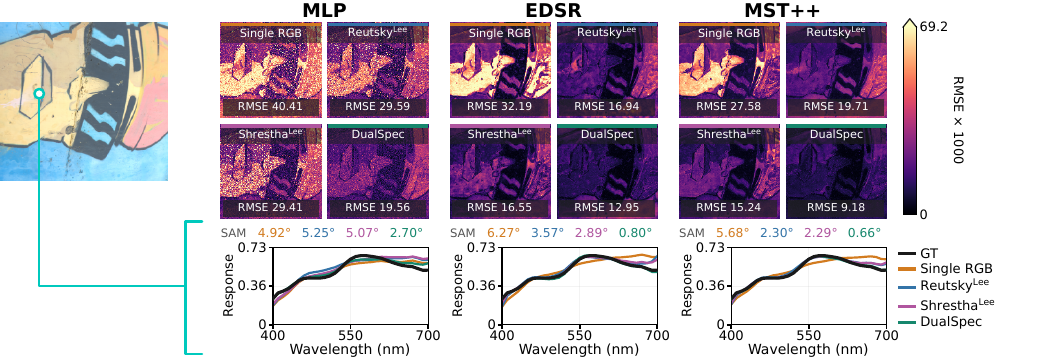}
\\\vspace{2mm}
\includegraphics[width=\textwidth]{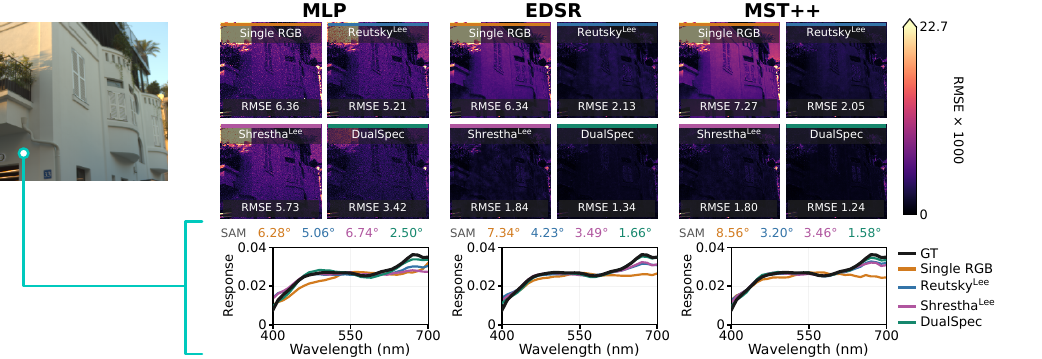}
\caption{Qualitative reconstruction on two Arad-1K test scenes. For each backbone, we compare single RGB, two stereo pairs selected from the Lee filter bank, and \ours{}. The figure shows spatial RMSE maps and reconstructed spectra at the marked pixels.\vspace{-2mm}}
\label{fig:qualitative_arad}
\end{figure*}

Table~\ref{tab:arad} shows the quantitative results on Arad-1K. A single learned attachment lowers mean SAM relative to unfiltered RGB for all three backbones, showing that reshaping the spectral sensitivities can make three measurements more informative. The complementary measurements provided by \ours{} yield a further gain. With the pixel-wise MLP, \ours{} reduces mean SAM to $4.73^\circ$, below both spatial backbones operating on unfiltered RGB ($5.28^\circ$ and $5.35^\circ$). Because the MLP processes each pixel independently without relying on semantic cues, this improvement can be fully attributed to the additional information provided by \ours{}. EDSR and MST++ further reduce the error ($2.02^\circ$ and $1.93^\circ$ against $4.73^\circ$), which suggests that spatial context helps to resolve part of the remaining ambiguity.

All stereo configurations provide six input channels, so their differences reflect the design of the transmittances rather than the number of measurements. At this equal channel count, \ours{} lowers mean SAM by 11--15\% relative to the best selected filter pair for each backbone. The gains extend to the median and tail statistics. For EDSR and MST++, the 95th-percentile SAM with \ours{} is even below the mean SAM of either single-camera configuration. 
For filter selection baselines, results show that the best catalog depends on the considered backbone: under the Reutsky~\etal{} criterion, Omega yields lower mean SAM for the MLP, while Lee does for EDSR and MST++. This further supports designing transmittances specifically for the network that will use them.

The spectra in~\Cref{fig:qualitative_arad} show what these gains look like at individual pixels: single RGB largely misses around 700\,nm, the filter selection baselines can sometimes recover part of this error, while \ours{} follows the measured spectrum more closely with all three backbones.

\begin{figure*}[t]
\centering
\includegraphics[width=\textwidth]{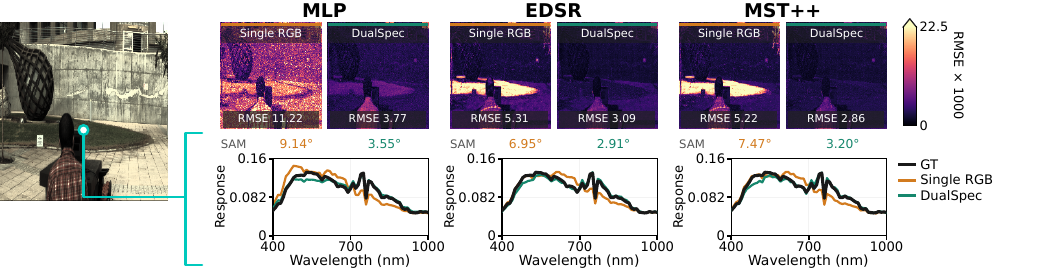}
\\ \vspace{2mm}
\includegraphics[width=\textwidth]{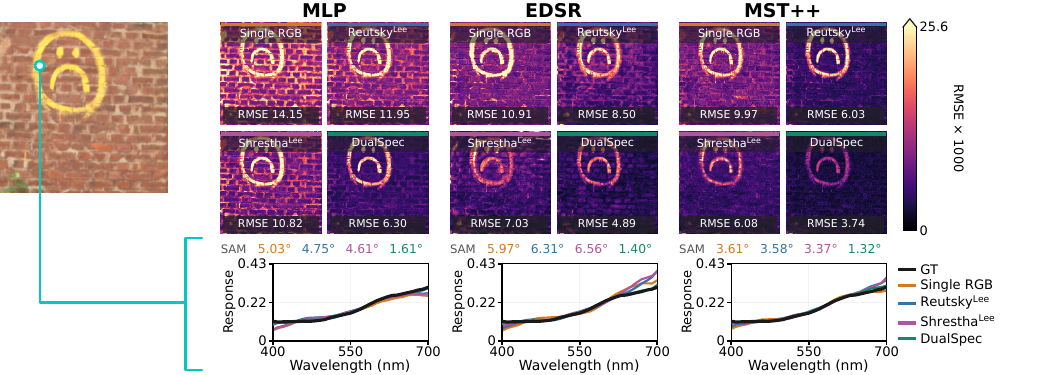}
\caption{Qualitative results on BGU-ICVL (top) and our displaced-view benchmark (bottom). The top example compares single-camera and DualSpec reconstruction over 400--1000\,nm. The bottom example compares single-camera, selected-filter stereo, and DualSpec reconstruction after registering the independently captured views.\vspace{-2mm}}
\label{fig:qualitative_vnir_real}
\end{figure*}

\paragraph{Extending the spectral range to near-infrared.}\label{sec:vnir}
Several applications require spectral information beyond the visible range~\cite{kim2025pixel,liang2022multimodal}. We therefore evaluate \ours{} over 400--1000\,nm, retraining both attachment transmittances and reconstruction networks for this wider interval on the BGU-ICVL dataset~\cite{arad2016sparse}. Commercial-filter stereo baselines are excluded because the available Lee and Omega filter data cover only the visible range.
Table~\ref{tab:vnir} reports the VNIR reconstruction results. For the pixel-wise MLP, a single learned attachment brings mean SAM from $8.53^\circ$ to $3.31^\circ$, showing that redistributing sensitivity across the wider band helps even with three channels. However, the optimized stereo pair cuts the remaining error by 29\% and, as for the results on Arad-1K, it  brings the pixel-wise model below both spatial models using unfiltered RGB in mean SAM. 
The extra measurements also help models that already exploit spatial context. For EDSR and MST++, \ours{} reduces both mean SAM and its 95th percentile relative to either single-camera input.
The top panel of~\Cref{fig:qualitative_vnir_real} shows a qualitative example: the stereo input helps all three backbones recover the marked spectrum's shape in both the visible and near-infrared ranges.

\begin{table}[t]
\caption{Cross-dataset spectral reconstruction on registered displaced-view pairs. The right view is the reference, and all methods are evaluated over the same valid region. Superscripts identify the filter bank used by each selection method. Bold values indicate the best result within each backbone and statistic, while blue shading identifies \ours{}.\vspace{-2mm}}\label{tab:misaligned}
\centering\scriptsize
\setlength{\tabcolsep}{1.8pt}
\resizebox{\columnwidth}{!}{%
\begin{tabular}{ll*{9}{c}}
\toprule
\multirow[c]{2}{*}[-2.5pt]{Backbone} &
\multirow[c]{2}{*}[-2.5pt]{Acquisition / selection} &
\multicolumn{3}{c}{SAM ($^\circ$) $\downarrow$} &
\multicolumn{3}{c}{RMSE $\times 1000\,\downarrow$} &
\multicolumn{3}{c}{PSNR $\uparrow$} \\
\cmidrule(lr){3-5}\cmidrule(lr){6-8}\cmidrule(l){9-11}
 & & Mean & Median & P95 & Mean & Median & P95 & Mean & Median & P05 \\
\midrule
\multirow[c]{7}{*}[-3.5pt]{MLP}
 & Single RGB & 8.57 & 8.94 & 12.28 & 42.1 & 42.3 & 67.8 & 27.96 & 27.49 & 23.38 \\
 & Single + learned filter & 7.88 & 8.25 & 10.77 & 35.7 & 35.9 & 59.4 & 29.54 & 28.89 & 24.53 \\
\noalign{\vskip 2pt}
\cdashline{2-11}
\noalign{\vskip 2pt}
 & Reutsky$^{\text{Lee}}$ & 7.26 & 7.43 & 10.42 & 34.3 & 36.2 & 52.8 & 29.77 & 28.85 & 25.55 \\
 & Shrestha$^{\text{Lee}}$ & 7.02 & 7.03 & 10.25 & 34.6 & 36.8 & 52.5 & 29.66 & 28.69 & 25.61 \\
 & Reutsky$^{\text{Omega}}$ & 6.50 & 6.71 & 9.53 & 31.9 & 32.0 & 48.2 & 30.49 & 29.90 & 26.37 \\
 & Shrestha$^{\text{Omega}}$ & 7.41 & 7.57 & 10.61 & 39.3 & 38.2 & 59.9 & 28.58 & 28.34 & 24.48 \\
\rowcolor{DualSpecBlue}
\cellcolor{white}
 & DualSpec (Ours) & \textbf{5.25} & \textbf{4.67} & \textbf{8.57} & \textbf{24.5} & \textbf{24.9} & \textbf{39.1} & \textbf{32.80} & \textbf{32.11} & \textbf{28.20} \\
\midrule
\multirow[c]{7}{*}[-3.5pt]{EDSR}
 & Single RGB & 6.79 & 6.98 & 9.20 & 34.5 & 35.7 & 56.2 & 29.76 & 28.94 & 25.01 \\
 & Single + learned filter & 8.31 & 9.24 & 10.60 & 39.1 & 40.3 & 68.0 & 29.22 & 27.99 & 23.40 \\
\noalign{\vskip 2pt}
\cdashline{2-11}
\noalign{\vskip 2pt}
 & Reutsky$^{\text{Lee}}$ & 5.20 & 5.15 & 7.69 & 26.0 & 27.5 & 41.3 & 32.59 & 31.31 & 27.68 \\
 & Shrestha$^{\text{Lee}}$ & 5.57 & 4.86 & 10.02 & 30.4 & 27.3 & 65.3 & 31.85 & 31.29 & 23.93 \\
 & Reutsky$^{\text{Omega}}$ & 4.76 & 4.39 & 8.49 & 25.4 & 24.3 & 47.3 & 33.04 & 32.31 & 26.69 \\
 & Shrestha$^{\text{Omega}}$ & 7.30 & 6.05 & 14.42 & 43.3 & 32.2 & 111.7 & 29.63 & 29.85 & 19.67 \\
\rowcolor{DualSpecBlue}
\cellcolor{white}
 & DualSpec (Ours) & \textbf{3.72} & \textbf{3.25} & \textbf{5.99} & \textbf{20.1} & \textbf{21.7} & \textbf{31.1} & \textbf{34.77} & \textbf{33.29} & \textbf{30.14} \\
\midrule
\multirow[c]{7}{*}[-3.5pt]{MST++}
 & Single RGB & 6.23 & 6.05 & 8.87 & 33.3 & 28.5 & 60.8 & 30.33 & 30.90 & 24.31 \\
 & Single + learned filter & 7.61 & 7.86 & 9.52 & 40.3 & 41.1 & 67.6 & 28.63 & 27.71 & 23.42 \\
\noalign{\vskip 2pt}
\cdashline{2-11}
\noalign{\vskip 2pt}
 & Reutsky$^{\text{Lee}}$ & 4.73 & 4.80 & 7.27 & 23.9 & 23.9 & 37.2 & 33.22 & 32.46 & 28.60 \\
 & Shrestha$^{\text{Lee}}$ & 5.53 & 4.89 & 10.61 & 31.4 & 25.2 & 75.8 & 32.13 & 31.99 & 22.58 \\
 & Reutsky$^{\text{Omega}}$ & 3.97 & 3.58 & 6.86 & 22.9 & 21.0 & 42.3 & 33.80 & 33.57 & 27.62 \\
 & Shrestha$^{\text{Omega}}$ & 6.53 & 5.65 & 12.38 & 41.0 & 29.7 & 100.0 & 29.81 & 30.60 & 20.99 \\
\rowcolor{DualSpecBlue}
\cellcolor{white}
 & DualSpec (Ours) & \textbf{3.48} & \textbf{3.29} & \textbf{5.70} & \textbf{18.2} & \textbf{19.4} & \textbf{28.5} & \textbf{35.73} & \textbf{34.26} & \textbf{30.93} \\
\bottomrule
\end{tabular}}\vspace{-2mm}
\end{table}

\paragraph{Displaced spectral views.}\label{sec:real}
We test whether the Arad-1K designs transfer when the two views come from different camera positions. 
To this end, we employ the displaced-view benchmark described in~\Cref{sec:data}.
We synthesize RGB inputs from independently captured misaligned spectral cubes using the acquisition model described in~\Cref{sec:camera} and apply the filters and networks optimized on Arad-1K without fine-tuning. For each stereo configuration, we estimate a homography from SIFT matches with RANSAC~\cite{lowe1999object,fischler1981random}. Single-camera methods use the right view alone, and all methods are evaluated over the same valid overlap.

Table~\ref{tab:misaligned} shows reconstruction accuracy on the registered displaced views. The single learned attachment, which helped on Arad-1K, raises mean SAM for both spatial backbones on these scenes. We attribute this to the ambiguity that remains with three channels, which the networks resolve with priors learned from the training scenes. The six complementary measurements of \ours{} reduce this reliance. The learned stereo pair also retains an advantage over the best selected stereo baselines for each backbone, lowering mean SAM by 12--22\%, and its gains extend to the upper end of the error distribution. The bottom panel of~\Cref{fig:qualitative_vnir_real} shows a qualitative example on the displaced-view benchmark, where the error maps and plotted spectra confirm the gains across all backbones.

On the 32 scenes of CAVE~\cite{yasuma2010generalized}, evaluated in the same cross-dataset setting, \ours{} lowers mean SAM by 18--29\% relative to the best selected pair (see supplementary material).
These results show that the acquisition gains persist also in cross-dataset evaluations and with real viewpoint misalignments.\vspace{-2mm}

\begin{table}[t]
\centering
\caption{Attachment parameterization on Arad-1K (mean SAM in degrees).
OD: optical-density bases factorized from the Lee or Omega catalog;
Gaussian: Eq.~\eqref{eq:basis} with spacing $\Delta_g$ and width $\sigma_g$ in nm;
$B$: number of bases.\vspace{-2mm}}
\label{tab:ablation}
\footnotesize
\setlength{\tabcolsep}{4.5pt}
{\begin{tabular}{lcccccc}
\toprule
\multirow{2.5}{*}{Attachment basis} &
\multirow{2.5}{*}{$\Delta_g$} &
\multirow{2.5}{*}{$\sigma_g$} &
\multirow{2.5}{*}{$B$} &
\multicolumn{3}{c}{Mean SAM ($^\circ$) $\downarrow$} \\
\cmidrule(lr){5-7}
& & & & MLP & EDSR & MST++ \\
\midrule
Best selected pair & -- & -- & -- & 5.33 & 2.39 & 2.28 \\
\noalign{\vskip 2pt}
\hdashline
\noalign{\vskip 2pt}
Lee OD & -- & -- & 8 & 5.40 & 2.54 & 2.14 \\
Omega OD & -- & -- & 12 & 4.85 & 2.27 & 2.11 \\
Gaussian & 40 & 40 & 8 & 4.87 & 2.13 & 2.06 \\
Gaussian & 20 & 40 & 16 & 4.80 & 2.08 & 2.09 \\
Gaussian & 20 & 20 & 16 & 4.84 & 2.10 & 2.09 \\
Gaussian & 20 & 10 & 16 & 4.95 & 2.06 & 2.04 \\
\rowcolor{DualSpecBlue}
Gaussian (\ours) & 10 & 10 & 31 & \textbf{4.73} & \textbf{2.02} & \textbf{1.93} \\
\bottomrule
\end{tabular}\vspace{-2mm}}
\end{table}

\paragraph{Choice of the attachment basis.}\label{sec:ablation}
Table~\ref{tab:ablation} compares the Gaussian basis of~\cref{eq:basis} with bases derived from the commercial catalogs, in which each attachment is a nonnegative combination of optical densities obtained by factorizing the Lee or Omega filters. Restricting the attachments to the spectral shapes of existing filters reduces accuracy: with EDSR and MST++, every Gaussian setting outperforms both catalog bases, and the Lee basis is the least accurate with every backbone. Within the Gaussian family, mean SAM varies by at most 0.22$^\circ$, 0.11$^\circ$, and 0.16$^\circ$ for the MLP, EDSR, and MST++, so the method is not sensitive to the exact choice of the basis. The configuration using $\Delta_g=\sigma_g=10$\,nm is the most accurate with every backbone, and therefore the selected one for the main experiments.\vspace{-2mm}

\begin{figure}[t]
\centering
\includegraphics[width=\columnwidth]{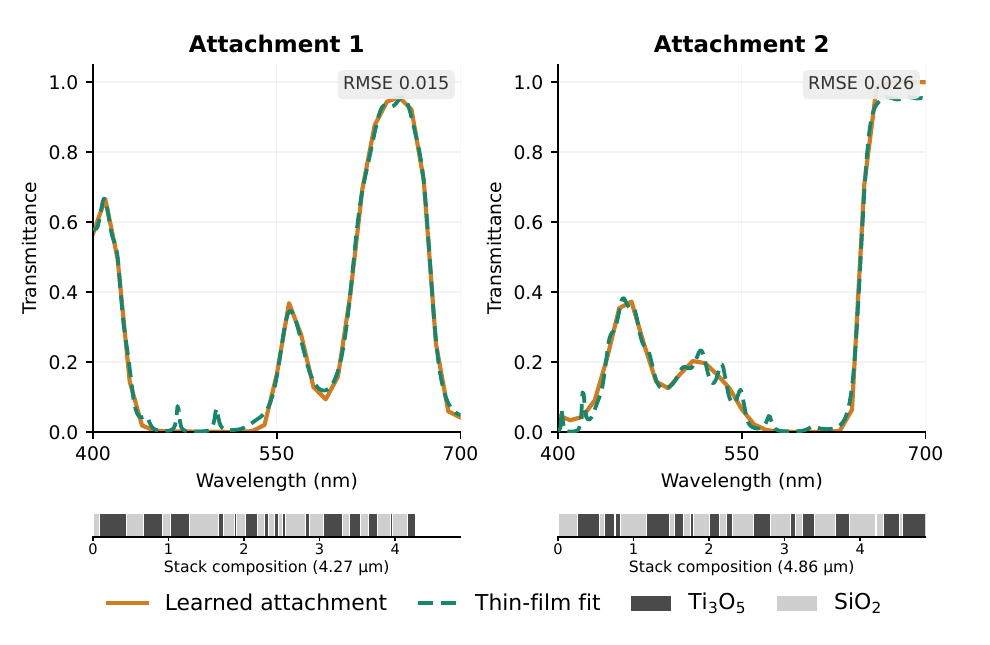}
\caption{Thirty-layer thin-film approximations of the two attachments learned with MST++. The curves compare the learned targets with their thin-film fitted transmittances. The bars below show the alternating $\mathrm{Ti_3O_5}$ and $\mathrm{SiO_2}$ layers to scale. The same thin-film approximation plots for the other backbones can be found in the supplementary material.\vspace{-2mm}}
\label{fig:physical_j30}
\end{figure}

\paragraph{Thin-film filter approximation.}\label{sec:analysis}
The basis parameterization of~\cref{eq:basis} guarantees smooth transmittances in the $(0,1)$ range. To assess whether the learned transmittances can be realized with conventional multilayer interference filters, we fit each attachment with a 30-layer stack of alternating $\mathrm{Ti_3O_5}$ and $\mathrm{SiO_2}$ layers on a glass substrate, using a standard thin-film optics model~\cite{macleod2010thin}. \Cref{fig:physical_j30} shows the fitted responses for MST++ on Arad-1K. Across backbones, the mean fitting RMSE ranges from 0.021 to 0.032, and the stack thicknesses range from 4.3 to 5.4\,\si{\micro\metre}.

\begin{table}[t]
\caption{Thin-film approximation on Arad-1K with $J=30$ layers. \emph{Thin film}: the learned attachments are replaced by the transmittances of the fitted stacks, with the unchanged network; \emph{+ fine-tuning}: the network is fine-tuned on the thin-film attachments. \emph{Att.\ RMSE}: thin-film against learned transmittance, mean of the two attachments; \emph{Thick.}: thickness of the thicker of the two stacks.\vspace{-2mm}}
\label{tab:thin_film_j30}
\centering\scriptsize
\setlength{\tabcolsep}{2.5pt}
\resizebox{\columnwidth}{!}{%
\begin{tabular}{ll*{5}{c}}
\toprule
Backbone & Attachments & Att.\ RMSE$\downarrow$ & Thick.\ (\si{\micro\metre}) & SAM ($^\circ$)$\downarrow$ & RMSE$\times$1000$\downarrow$ & PSNR$\uparrow$ \\
\midrule
& Reutsky$^{\text{Omega}}$ & -- & -- & 5.33 & 13.76 & 39.60 \\
\cellcolor{white}
 & Learned (\ours) & -- & -- & 4.73 & 10.93 & 41.59 \\
\noalign{\vskip 1pt}
\cdashline{2-7}
\noalign{\vskip 1pt}
 & Thin film, $J=30$ & \multirow[c]{2}{*}{0.032} & \multirow[c]{2}{*}{5.09} & 5.14 & 12.69 & 40.48 \\
\multirow[c]{-4}{*}{MLP} & \quad + fine-tuning & & & 4.87 & 11.37 & 41.26 \\
\midrule
 & Reutsky$^{\text{Lee}}$ & -- & -- & 2.39 & 6.37 & 46.36 \\
\cellcolor{white}
 & Learned (\ours) & -- & -- & 2.02 & 5.30 & 47.98 \\
\noalign{\vskip 1pt}
\cdashline{2-7}
\noalign{\vskip 1pt}
 & Thin film, $J=30$ & \multirow[c]{2}{*}{0.021} & \multirow[c]{2}{*}{5.35} & 2.43 & 6.77 & 45.80 \\
\multirow[c]{-4}{*}{EDSR} & \quad + fine-tuning & & & 2.03 & 5.32 & 47.95 \\
\midrule
 & Reutsky$^{\text{Lee}}$ & -- & -- & 2.28 & 5.93 & 46.51 \\
\cellcolor{white}
 & Learned (\ours) & -- & -- & 1.93 & 4.95 & 48.17 \\
\noalign{\vskip 1pt}
\cdashline{2-7}
\noalign{\vskip 1pt}
 & Thin film, $J=30$ & \multirow[c]{2}{*}{0.021} & \multirow[c]{2}{*}{4.86} & 2.50 & 7.09 & 45.11 \\
\multirow[c]{-4}{*}{MST++} & \quad + fine-tuning & & & 1.94 & 4.98 & 48.14 \\
\bottomrule
\end{tabular}}
% \vspace{-2mm}
\end{table}

After fabrication, the network can be fine-tuned on measured filter transmittances. Here, we fine-tune each network on the fitted thin-film transmittances for 60 epochs at a learning rate of $10^{-4}$, starting from its original weights. As shown in~\Cref{tab:thin_film_j30}, with frozen weights, the fitted transmittances raise mean SAM by $\sim$20\% for EDSR and $\sim$29\% for MST++. With fine-tuning, all three networks outperform their best catalog pairs and come within $0.01^\circ$ (EDSR and MST++) or $0.14^\circ$ (MLP) of the mean SAM achieved with the optimized transmittances.
Together, these results show that the performance gains of \ours{} largely survive thin-film filter approximation, with a brief fine-tuning step recovering most of the loss.\vspace{-4mm}

\paragraph{Limitations.}
Our current design uses identical sensors to isolate the contribution of the learned attachments and simplify stereo acquisition. Employing different sensors could provide additional spectral diversity, but would require accounting for differences in quantum efficiency, exposure, synchronization, and field of view.

\section{Conclusion}\label{sec:conclusion}

We presented \ours{}, a stereo spectral-reconstruction framework that addresses the metameric ambiguity of RGB measurements by jointly optimizing two passive optical attachments and the reconstruction network. Across three reconstruction families and both visible and VNIR ranges, the complementary stereo measurements consistently reduce spectral error relative to single-camera acquisition and to stereo pairs selected from commercial filter banks. Cross-dataset evaluation on independently captured displaced views further demonstrates that the learned acquisition transfers without fine-tuning, while experiments on thin-film fits show that the designs can be manufactured with conventional multilayer coatings and recover the learned accuracy after brief fine-tuning. Overall, our results show that advancing spectral recovery is not only a matter of building more complex reconstruction models, but of designing measurements that make spectra distinguishable in the first place.

\section*{LLM Usage Statement}

OpenAI GPT-5.6, GPT-6, GPT-6.1, and Anthropic Claude Opus 5.5 were used to assist with manuscript preparation, specifically for rephrasing and grammar checking, and with writing scripts that visualize existing experimental outputs. These tools were not used to generate data, conduct experiments, derive results, or write the implementation code for the proposed method. All methodological decisions, analyses, visualizations, and final wording were reviewed and approved by the authors, who take full responsibility for the submitted work.
\section*{Acknowledgments}

Computational resources provided by hpc-ReGAInS@DISCo, which is a MUR Department of Excellence Project within the Department of Informatics, Systems and Communication at the University of Milan-Bicocca (https://www.disco.unimib.it)
\hypersetup{pageanchor=false}
% \clearpage
% \setcounter{page}{1}
% \maketitlesupplementary
\section*{Supplementary Material}

\setcounter{section}{0}
\setcounter{figure}{0}
\setcounter{table}{0}
\setcounter{equation}{0}
\renewcommand{\thesection}{\Alph{section}}
\renewcommand{\thefigure}{S\arabic{figure}}
\renewcommand{\thetable}{S\arabic{table}}
\renewcommand{\theequation}{S\arabic{equation}}
\renewcommand{\theHsection}{supp.\arabic{section}}
\renewcommand{\theHfigure}{supp.\arabic{figure}}
\renewcommand{\theHtable}{supp.\arabic{table}}
\renewcommand{\theHequation}{supp.\arabic{equation}}

We provide additional material that complements our main submission. The supplementary document includes:
\begin{itemize}
    \item details of the commercial filter catalogs and on filter selection baselines (Section~\ref{sec:supp_filter_selection});
    \item the attachments learned on Arad-1K and BGU-ICVL with every backbone (Section~\ref{sec:supp_all_filters});
    % \item additional analysis of the learned VNIR responses on BGU-ICVL (Section~\ref{sec:supp_bgu_vnir});
    \item the theoretical foundation for thin-film filters, details on the fitting and fine-tuning procedure, and the results of thin-film fitting on BGU-ICVL (Section~\ref{sec:supp_thinfilm});
    \item acquisition and evaluation details for the displaced-view benchmark, results with the left view as the reference, and results without misalignment (Section~\ref{sec:supp_displaced_view});
    \item cross-dataset evaluation on CAVE dataset (Section~\ref{sec:supp_cave});
    \item additional qualitative results on Arad-1K, BGU-ICVL, and the displaced-view benchmark (Section~\ref{sec:supp_qualitative}).
\end{itemize}

\section{Commercial Filter Catalogs}
\label{sec:supp_filter_selection}

\paragraph{Candidate catalogs.}
We construct two banks from manufacturer-published visible-range transmittance curves. The Lee bank is derived from 219 lighting filters sampled from 405 to 700\,nm at 5\,nm intervals~\cite{lee_filters_catalog}. We remove nine diffusion filters because their spatial scattering cannot be represented by a scalar spectral transmittance, leaving 210 filters. The Omega Optical catalog is composed by interference and absorption-glass filters having sharper bandpass, longpass, shortpass, notch, and multiband responses~\cite{omega_filters_catalog}. Of 474 cataloged filters, 470 provide numerical curves. Curves expressed as optical density are converted using
\begin{equation}
    T(\lambda)=10^{-\mathrm{OD}(\lambda)}.
\end{equation}
We retain filters whose measurements cover the full visible interval (400--700\,nm) and reach at least 1\% transmittance. We exclude mirrors, uncoated blanks, duplicated curves, and filters characterized at nonzero incidence angles. This yields 136 Omega filters. ~\Cref{fig:supp_filter_banks} shows representative responses from both resulting filter banks.
Curves are converted to the 31 spectral bands of the visible-range experiments (400--700\,nm) by linear interpolation and averaging over each 10\,nm band.

\begin{figure}[t]
\centering
\includegraphics[width=\columnwidth]{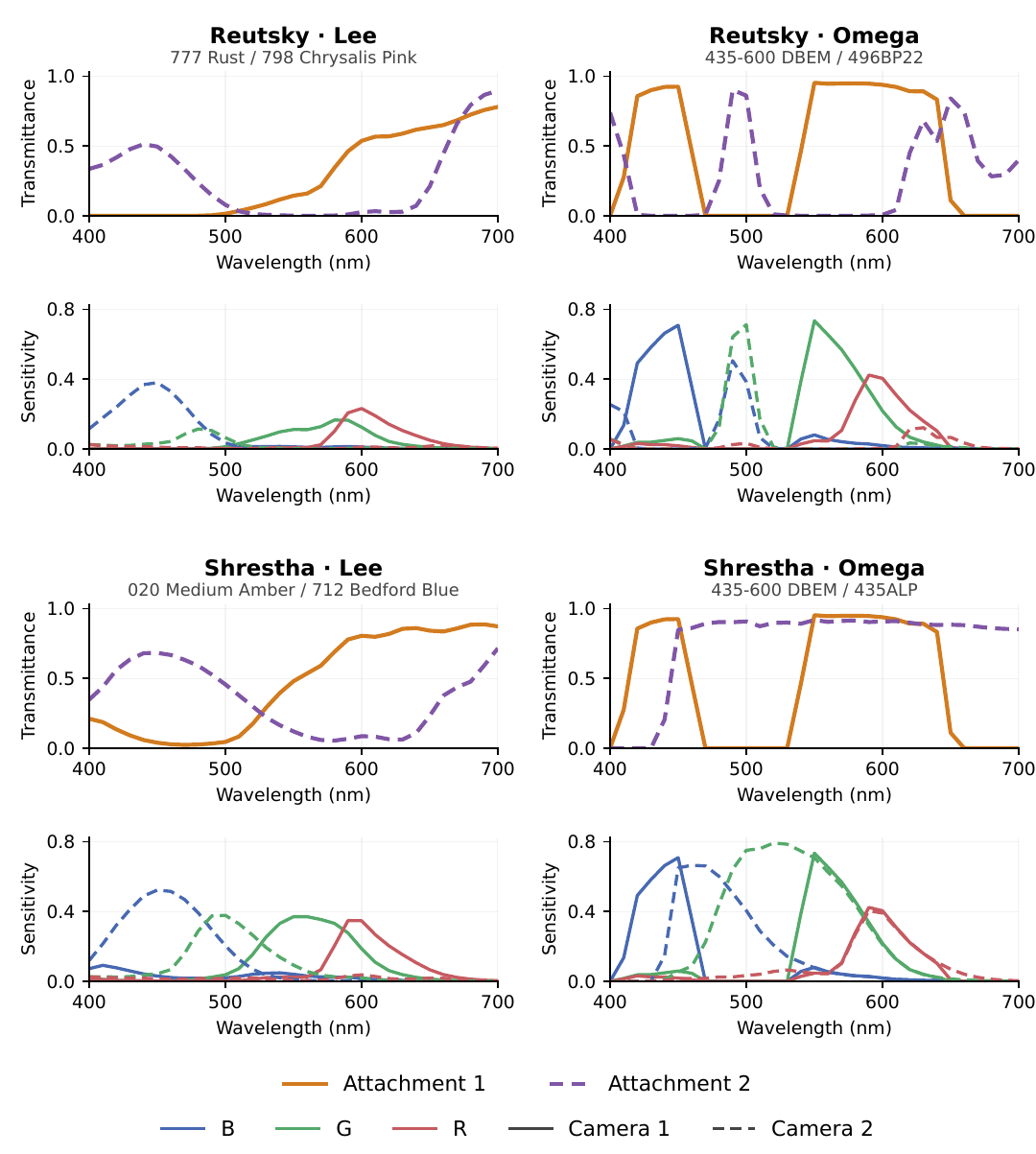}
\caption{Filter pairs selected from the Lee and Omega Optical banks by the adapted Reutsky and Shrestha criteria: transmittances (top) and resulting effective sensitivities of the two cameras (bottom) on Arad-1K. Solid and dashed curves correspond to the first and second cameras, respectively.}
\label{fig:supp_selected_filters}
\end{figure}

\begin{figure*}[t]
% \centering
\begin{tabular}{@{}c@{\hspace{0.05\textwidth}}c@{}}
  \small Lee & \small Omega Optical \\[4pt]
  \includegraphics[width=0.45\textwidth]
    {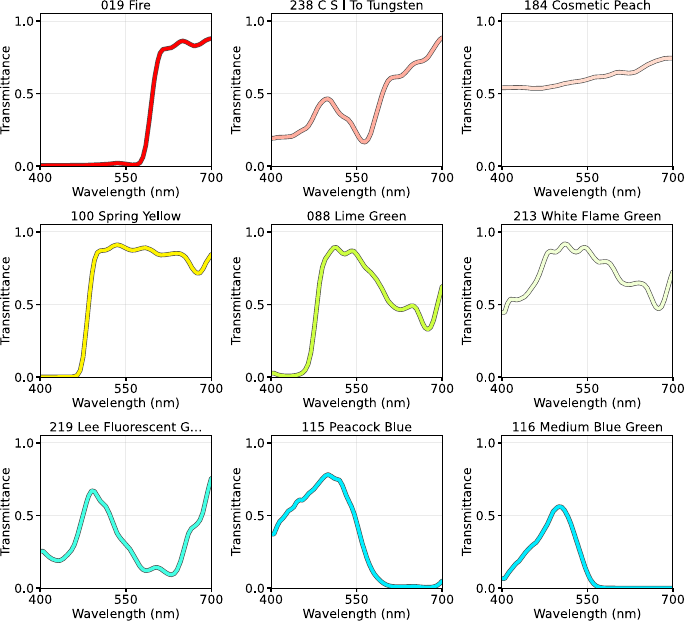}
  &
  \includegraphics[width=0.45\textwidth]
    {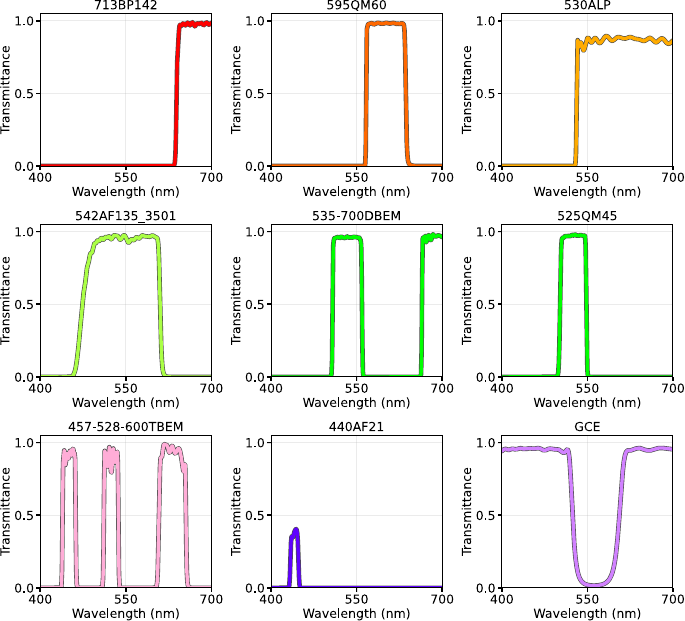}
\end{tabular}
\caption{Representative transmittance curves from the retained Lee (left) and Omega Optical (right) filter banks. Lee gels predominantly provide broad absorptive responses, whereas the Omega catalog also contains sharper interference-filter responses such as bandpass, notch, and multiband designs.}
\label{fig:supp_filter_banks}
\end{figure*}

For each bank, we also allow a clear response so that one camera may remain unfiltered. Since the two sensors are identical, filter pairs are unordered, and pairs containing the same response twice are excluded. A bank of $F$ filters therefore gives
\begin{equation}
    N_{\mathrm{pair}}=\binom{F+1}{2}=\frac{F(F+1)}{2}
\end{equation}
candidate pairs. This produces 22,155 Lee pairs and 9,316 Omega pairs. 

\paragraph{Adapted Shrestha selection.}
We adapt the exhaustive stereo-filter search of Shrestha \etal~\cite{shrestha2011multispectral} to two identical RGB cameras. We screen each filter independently, requiring its largest channel transmission factor to be at least 40\% and every channel transmission factor to be at least 10\% of that maximum. We evaluate all feasible pairs, including pairs with an unfiltered response. For each pair, we add Gaussian noise with a standard deviation of 2\% of the signal, normalize the six responses by the pair's largest white response, and quantize them to 12 bits.

Shrestha \etal trained on 63 ColorChecker DC patches selected by linear distance maximization~\cite{pellegri2004training}.
For each feasible pair, we use the same selection approach to retain 63 of 2,000 spectra randomly drawn from the Arad-1K training split~\cite{arad2022ntire}. We use these spectra and their simulated responses to fit a Tikhonov-regularized linear regressor~\cite{tikhonov1963regularization}.

For each filter catalog, we select the pair with the lowest mean CIELAB color difference $\Delta E^*_{ab}$ across 29,070 spectra subsampled from the Arad-1K validation split.
We compute color differences under standard illuminant D50 using the CIE 1964 $10^\circ$ standard observer~\cite{isoCie11664-4,isoCie11664-2,isoCie11664-1}.
The selected pairs are LEE 020 Medium Amber with 712 Bedford Blue and Omega 435-600 DBEM with 435ALP.

\paragraph{Adapted Reutsky selection.}
Following Reutsky \etal~\cite{reutsky2025modulate}, we rank filter pairs by the spectral uncertainty remaining after observing their noisy responses. For a pair $P$, we compute 
\begin{equation}
v(P) =
\mathbb{E}_{\mathbf{x}}\!\left[
  \operatorname{tr}\!\left(
    \operatorname{Cov}(\mathbf{y}\mid\mathbf{x})
  \right)
\right],
\label{eq}
\end{equation}
where $\mathbf{y}$ is the spectrum and $\mathbf{x}$ is its noisy six-channel measurement for $P$. A lower score indicates a more informative pair.

We follow Reutsky \etal's prior construction by subsampling images with a 29-pixel stride and clustering the resulting spectra into 1,024 groups using $k$-means. Our prior uses 232,254 spectra from the Arad-1K training split~\cite{arad2022ntire}, with each cluster center weighted by the fraction of spectra assigned to it. We simulate the six-channel measurements using the camera model described in the main manuscript. 
% Our two identical cameras allow us to evaluate unordered filter pairs, with a clear response as a candidate for an unfiltered view.
Our setup involves two identical cameras, therefore we evaluate unordered filter pairs.

We estimate Eq.~\eqref{eq} in three Monte Carlo stages, using the same random draws for all pairs within each stage. We score all pairs with 4,096 samples, rescore the top 2,000 with 65,536 samples, and rescore the top 50 from that stage with $2^{20}$ samples. 
The lowest-scoring pairs in the LEE and Omega catalogs are LEE 777 Rust with 798 Chrysalis Pink and Omega 435-600 DBEM with 496BP22 (1.25-inch NPB), respectively.\vspace{-4mm}

\paragraph{Selected pairs.}
~\Cref{fig:supp_selected_filters} shows the four selected pairs and the resulting effective sensitivities. The criterion of Reutsky \etal selects filters with complementary passbands, so that each camera mainly measures the part of the spectrum attenuated by the other, whereas the feasibility screen of Shrestha \etal, which discards dark filters, leads to brighter pairs with broader passbands.

\section{Learned Attachments on Arad-1K and BGU-ICVL}
\label{sec:supp_all_filters}

\begin{figure}[t]
\centering
\includegraphics[width=\columnwidth]{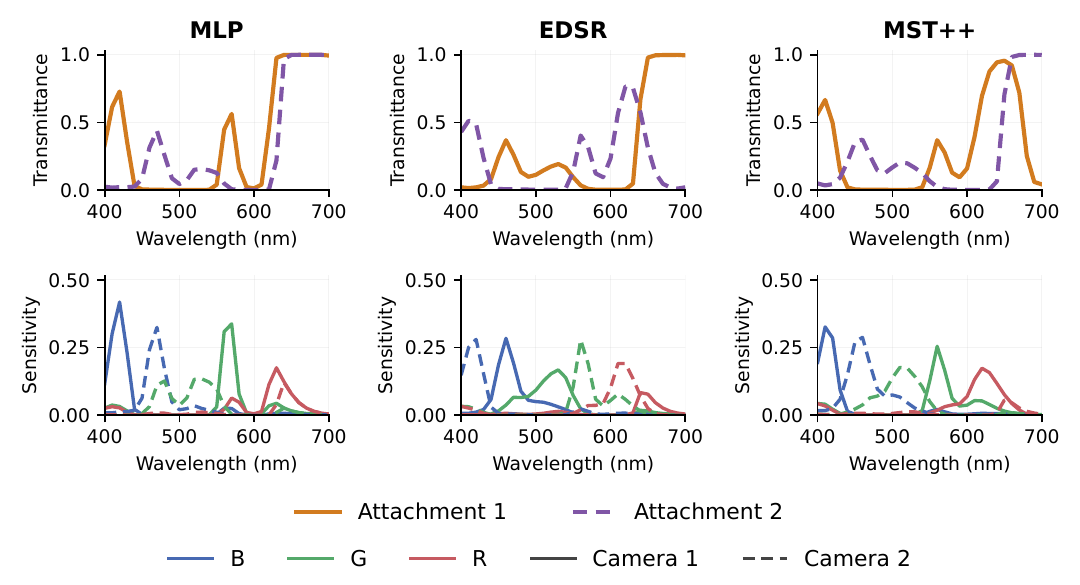}
\caption{Learned attachment transmittances (top) and resulting effective RGB sensitivities (bottom) for the MLP, EDSR, and MST++ reconstructors. Transmittances are absolute (the fraction of light passed), and sensitivities include them, so lower curves mean less collected light. Solid and dashed curves correspond to the first and second cameras, respectively.}
\label{fig:filters}
\end{figure}

\begin{figure}[t]
\centering
\includegraphics[width=\columnwidth]{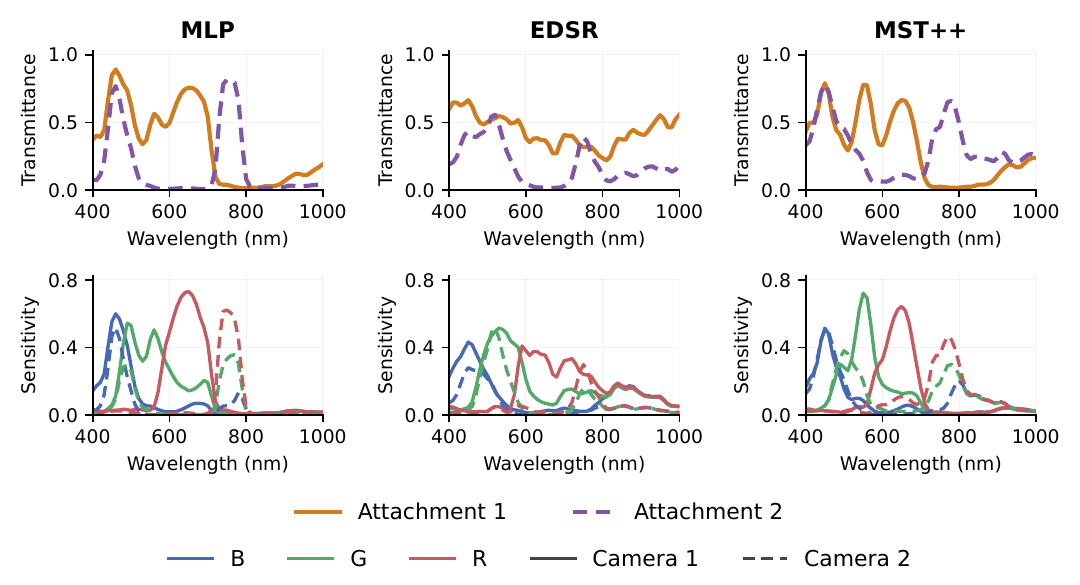}
\caption{Learned attachment transmittances (top) and resulting effective RGB sensitivities (bottom) for VNIR reconstruction on BGU-ICVL. Columns correspond to the MLP, EDSR, and MST++ reconstructors. Solid and dashed sensitivity curves denote the first and second cameras, respectively.}
\label{fig:supp_bgu_filters}
\end{figure}

Figures~\ref{fig:filters} and~\ref{fig:supp_bgu_filters} show the attachment transmittances learned by \ours{} and the resulting effective sensor sensitivities for Arad-1K and BGU-ICVL, respectively. 
The Arad-1K figure compares MLP, EDSR, and MST++, complementing Figure~2 of the main manuscript, which shows only MST++ designs. The BGU-ICVL figure shows results across the 400--1000\,nm visible and near-infrared (VNIR) range, providing a visual counterpart to the quantitative results in Table~2 of the main manuscript. The attachment pairs differ across backbones, reflecting the joint optimization of the optics and each reconstruction network.

\section{Thin-Film Realization}
\label{sec:supp_thinfilm}

\paragraph{Characteristic-matrix model.}
Following standard thin-film optics~\cite{macleod2010thin}, we model each attachment as a $J=30$ layer interference filter at normal incidence. The layers alternate between $\mathrm{Ti_3O_5}$ and $\mathrm{SiO_2}$, with refractive indices $n_H=2.40$ and $n_L=1.45$, respectively, and are deposited on a glass substrate with refractive index $n_s=1.52$. The incident medium is air, with $n_0=1$, and each layer thickness is constrained to $10\leq d_j\leq400$\,nm.

For layer $j$, with refractive index $n_j$ and thickness $d_j$, the phase delay and characteristic matrix are
\begin{equation}
    \delta_j(\lambda)
    =
    \frac{2\pi n_jd_j}{\lambda},
    \qquad
    \mathbf M_j(\lambda)
    =
    \begin{pmatrix}
        \cos\delta_j &
        \dfrac{\mathrm i}{n_j}\sin\delta_j\\[2pt]
        \mathrm i n_j\sin\delta_j &
        \cos\delta_j
    \end{pmatrix}.
    \label{eq:supp_layer_matrix}
\end{equation}
The characteristic matrix of the complete stack is obtained as
\begin{equation}
    \begin{pmatrix}B\\C\end{pmatrix}
    =
    \left(\prod_{j=1}^{J}\mathbf M_j\right)
    \begin{pmatrix}1\\n_s\end{pmatrix},
\end{equation}
from which its transmittance is
\begin{equation}
    T(\lambda;\mathbf d)
    =
    \frac{4n_0n_s}
         {(B+C)(B+C)^\ast},
    \label{eq:supp_stack_transmittance}
\end{equation}
where $\mathbf d=(d_1,\ldots,d_J)$ contains the layer thicknesses and $(\cdot)^\ast$ denotes complex conjugation. 
$T$ is first evaluated on a wavelength grid with 1\,nm spacing and then resampled to match the 10\,nm spacing of the other spectral data used in the experiments.

\begin{figure*}[t]
\centering
\includegraphics[width=0.325\textwidth]{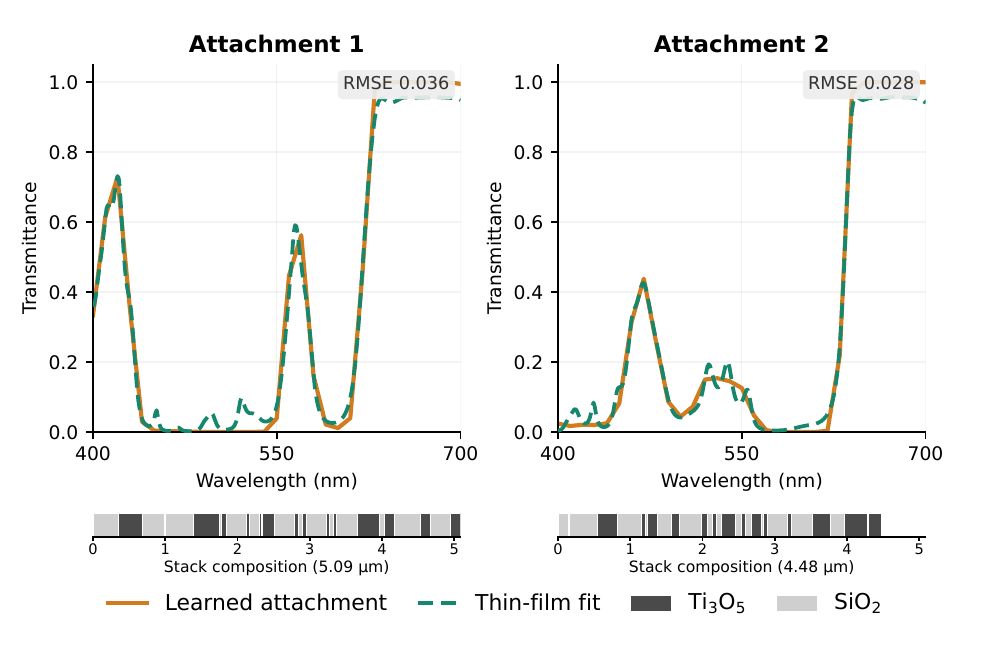}\hfill
\includegraphics[width=0.325\textwidth]{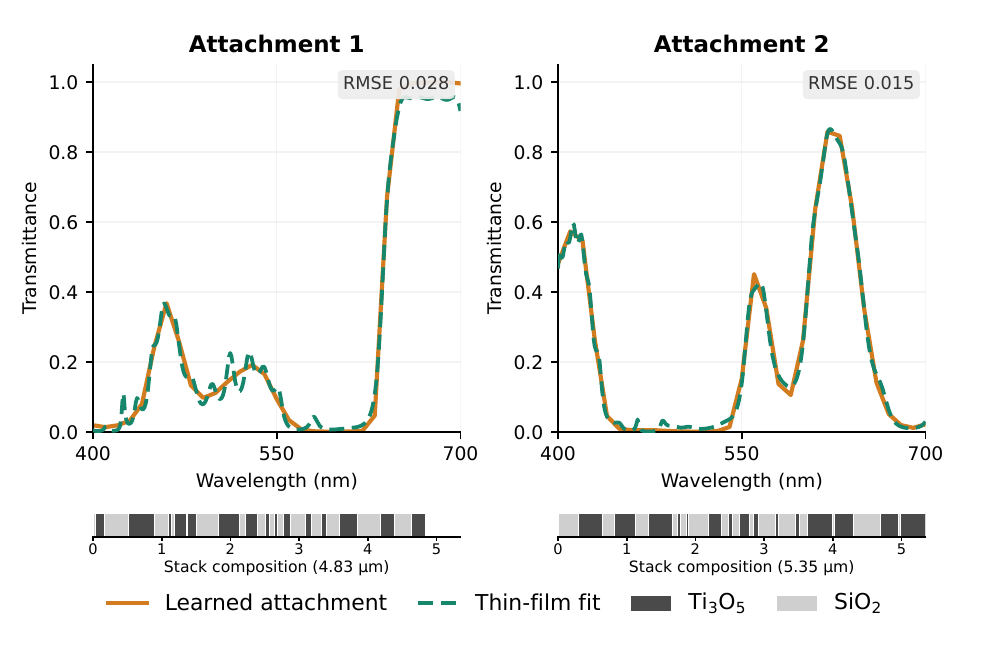}\hfill
\includegraphics[width=0.325\textwidth]{figures/supplementary_figures/thinfilm_arad_j30_mstpp.pdf}
\caption{Complete $J=30$ thin-film realizations on Arad-1K for MLP (left), EDSR (center), and MST++ (right). Each panel compares the learned and fitted transmittances for both attachments and shows the alternating $\mathrm{Ti_3O_5}$/$\mathrm{SiO_2}$ stacks to scale.}
\label{fig:supp_thinfilm_arad}
\end{figure*}

\begin{figure*}[t]
\centering
\includegraphics[width=0.325\textwidth]{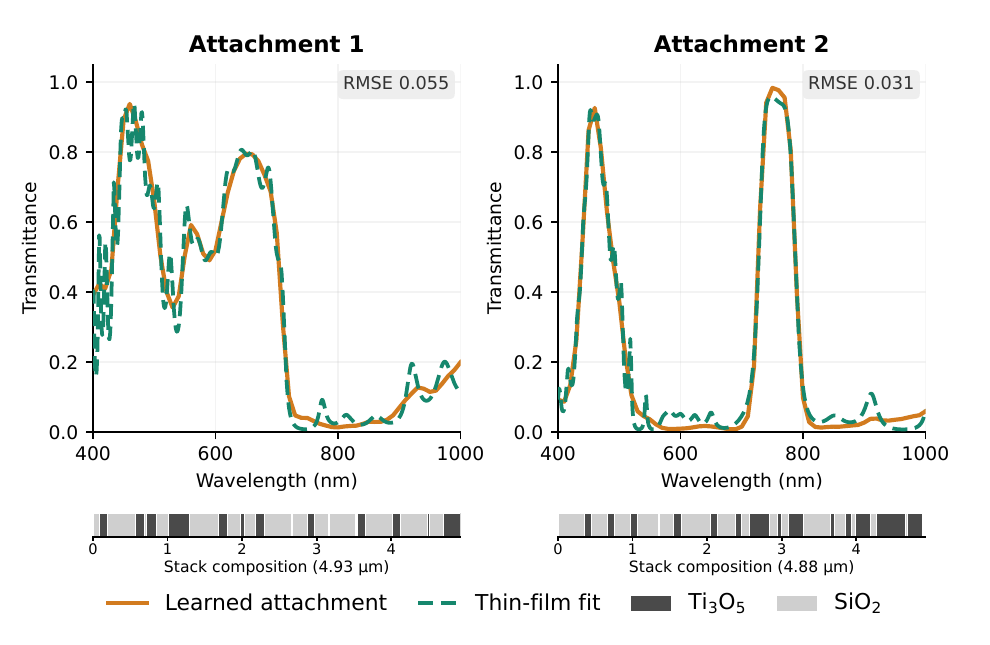}\hfill
\includegraphics[width=0.325\textwidth]{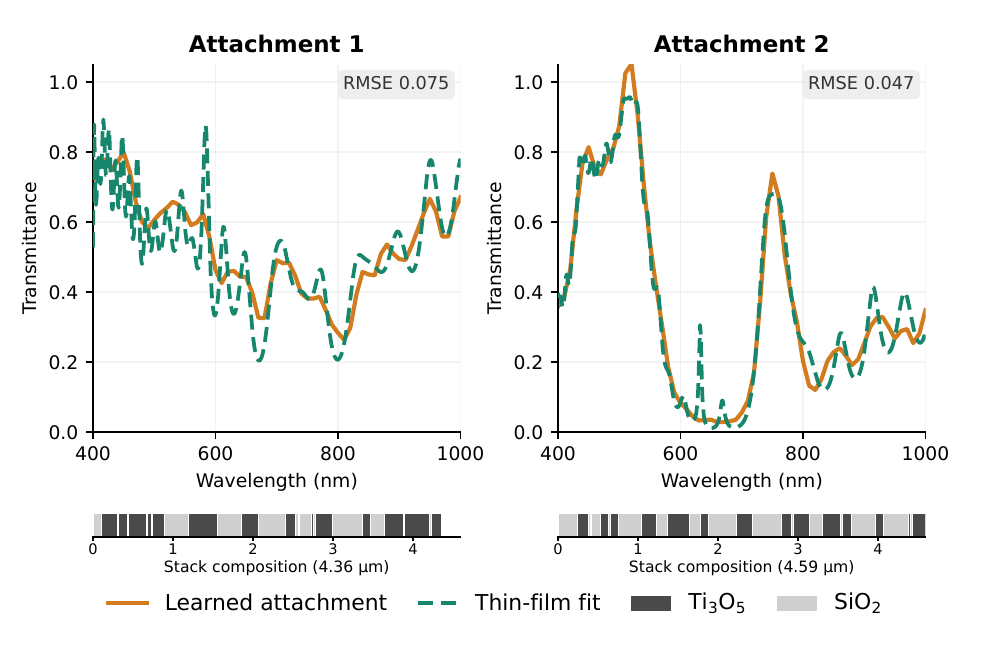}\hfill
\includegraphics[width=0.325\textwidth]{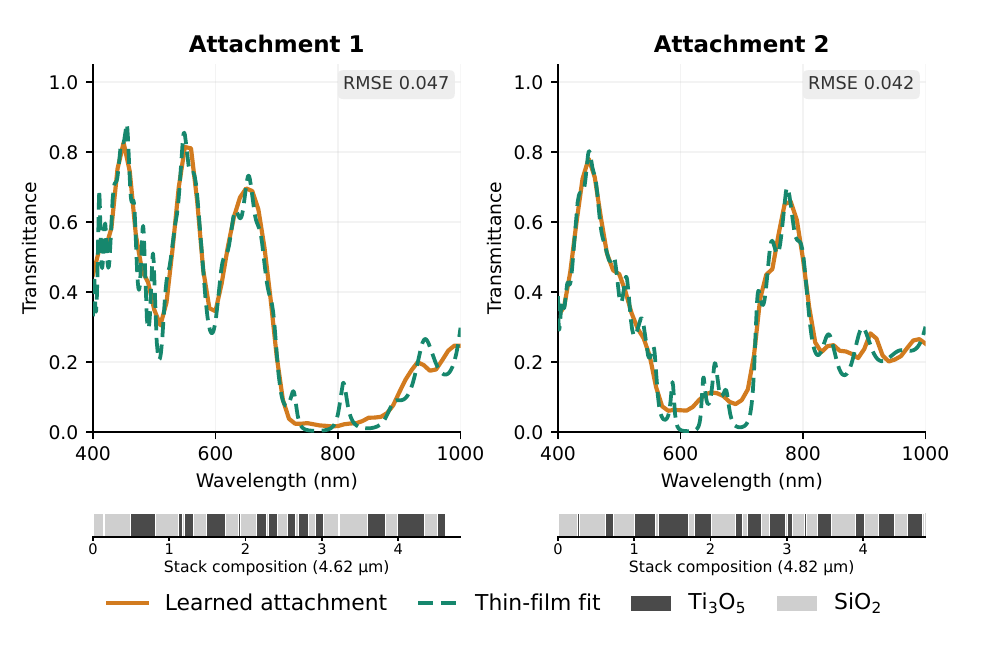}
\caption{Complete $J=30$ thin-film realizations over 400--1000\,nm on BGU-ICVL for MLP (left), EDSR (center), and MST++ (right). The wider spectral interval produces larger fitting errors than the visible-only Arad-1K designs.}
\label{fig:supp_thinfilm_bgu}
\end{figure*}

\paragraph{Thin-film fitting.}
We linearly interpolate each learned attachment $a_j$ onto a 1\,nm grid and fit the layer thicknesses by minimizing the mean-squared error between $sT(\lambda;\mathbf d)$ and $a_j(\lambda)$. The neutral-density factor $s\leq 1$ accounts for attenuation and is obtained in closed form by least squares. For each layer ordering ($\mathrm{Ti_3O_5}$-first and $\mathrm{SiO_2}$-first), we run 3,000 Adam steps~\cite{kingma2015adam} from each of 1,024 random initializations, refine the eight best candidates with bounded L-BFGS-B~\cite{byrd1995limited}, and retain the best design across both orderings.
We then resample its transmittance to 10\,nm steps and use it in the camera model.  Figures~\ref{fig:supp_thinfilm_arad} and~\ref{fig:supp_thinfilm_bgu} show the fitted transmittances and layer stacks for Arad-1K and BGU-ICVL, respectively.
For each backbone, we freeze the fitted attachments and fine-tune the trained reconstruction network for 60 epochs at a learning rate of $10^{-4}$, using the same experimental setup used in the second training stage described in the main manuscript.
Similarly to the results reported in the main manuscript for Arad-1K (Table~5), we report in \Cref{tab:supp_thinfilm_bgu} the results on BGU-ICVL. The wider spectral range makes the learned transmittances harder to reproduce with the same 30 layers.
Using the fitted filters with the original network weights increases mean SAM for all three backbones. Fine-tuning with the fitted filters recovers the reconstruction accuracy achieved with the learned attachments.

\section{Displaced-View Benchmark}
\label{sec:supp_displaced_view}

\begin{figure*}[t]
\centering
\includegraphics[width=\textwidth]{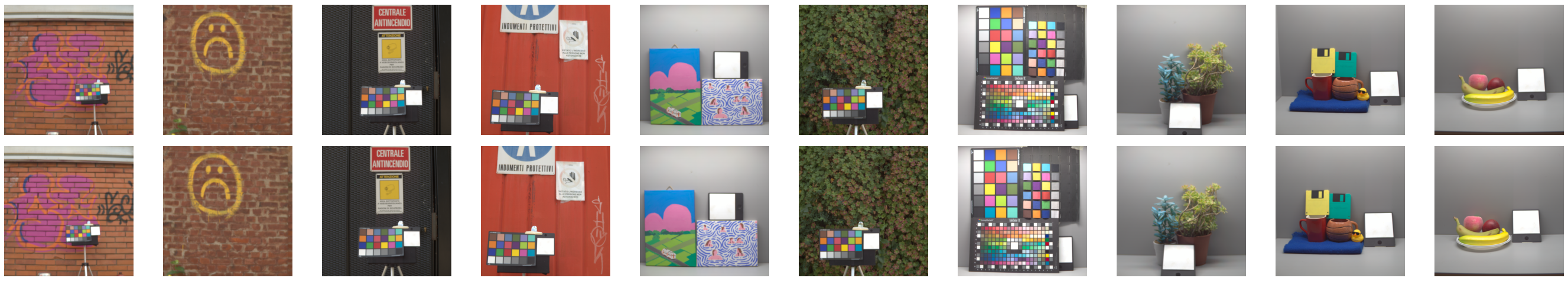}
\caption{The ten independently acquired spectral-image pairs in the displaced-view benchmark. Each L/R pair was captured with a 2\,cm horizontal camera displacement. Five scenes are staged indoor arrangements, the remaining five are outdoor scenes.}
\label{fig:supp_real_scenes}
\end{figure*}

\paragraph{Acquisition setup.}
We acquired ten pairs of spectral images using a Specim IQ hyperspectral camera~\cite{specimiq}. The camera was mounted on a bracket that allowed it to slide horizontally by a fixed displacement. For every pair, two independent acquisitions were made with a 2\,cm baseline between viewpoints, which is representative of close-range commercial stereo geometries~\cite{luxonis-oakdsr}.
The benchmark contains five staged indoor scenes and five outdoor scenes. 
~\Cref{fig:supp_real_scenes} shows both acquired views for all ten scenes.
The indoor subset includes color targets, fruit, paintings, plants, and arrangements of everyday objects. The outdoor subset includes vegetation, signs, brickwork, painted surfaces, and other textured structures.

\begin{table}[t]
\caption{Thin-film approximation of the learned attachments on BGU-ICVL over 400--1000\,nm with $J=30$ layers. \emph{Thin film}: the learned attachments are replaced by the transmittances of the fitted thin-film stacks, with the network left unchanged. \emph{+ fine-tuning}: the network is additionally fine-tuned on the thin-film attachments. \emph{Att.\ RMSE}: RMSE between the thin-film and learned transmittances, averaged over the two attachments. \emph{Thick.}: thickness of the thicker of the two stacks.}
\label{tab:supp_thinfilm_bgu}
\centering\scriptsize
\setlength{\tabcolsep}{2.5pt}
\resizebox{\columnwidth}{!}{%
\begin{tabular}{ll*{5}{c}}
\toprule
SSR & Attachments & Att.\ RMSE$\downarrow$ & Thick.\ (\si{\micro\metre}) & SAM ($^\circ$)$\downarrow$ & RMSE$\times$1000$\downarrow$ & PSNR$\uparrow$ \\
\midrule
\cellcolor{white}
\multirow[c]{3}{*}{MLP} & Learned (\ours) & -- & -- & 2.35 & 6.93 & 40.80 \\
\noalign{\vskip 1pt}
\cdashline{2-7}
\noalign{\vskip 1pt}
 & Thin film, $J=30$ & \multirow[c]{2}{*}{0.043} & \multirow[c]{2}{*}{4.93} & 2.46 & 7.37 & 40.28 \\
 & \quad + fine-tuning & & & 2.36 & 6.92 & 40.82 \\
\midrule
\cellcolor{white}
\multirow[c]{3}{*}{EDSR} & Learned (\ours) & -- & -- & 1.75 & 5.16 & 43.36 \\
\noalign{\vskip 1pt}
\cdashline{2-7}
\noalign{\vskip 1pt}
 & Thin film, $J=30$ & \multirow[c]{2}{*}{0.061} & \multirow[c]{2}{*}{4.59} & 1.88 & 5.84 & 42.13 \\
 & \quad + fine-tuning & & & 1.73 & 5.09 & 43.51 \\
\midrule
\cellcolor{white}
\multirow[c]{3}{*}{MST++} & Learned (\ours) & -- & -- & 1.74 & 5.28 & 43.13 \\
\noalign{\vskip 1pt}
\cdashline{2-7}
\noalign{\vskip 1pt}
 & Thin film, $J=30$ & \multirow[c]{2}{*}{0.044} & \multirow[c]{2}{*}{4.82} & 1.87 & 5.78 & 42.30 \\
 & \quad + fine-tuning & & & 1.71 & 5.18 & 43.26 \\
\bottomrule
\end{tabular}}
\end{table}

\begin{table}[t]

\caption{Cross-dataset spectral reconstruction on registered displaced-view pairs. The left view is the reference, and all methods are evaluated over the same valid region. Superscripts identify the filter bank used by each selection method. Bold values indicate the best result within each backbone and statistic, while blue shading identifies \ours{}.}\label{tab:supp_real_refL_noisy}
\centering\scriptsize
\setlength{\tabcolsep}{1.8pt}
\resizebox{\columnwidth}{!}{%
\begin{tabular}{ll*{9}{c}}
\toprule
\multirow[c]{2}{*}[-2.5pt]{SSR} &
\multirow[c]{2}{*}[-2.5pt]{Acquisition / selection} &
\multicolumn{3}{c}{SAM ($^\circ$) $\downarrow$} &
\multicolumn{3}{c}{RMSE $\times 1000\,\downarrow$} &
\multicolumn{3}{c}{PSNR $\uparrow$} \\
\cmidrule(lr){3-5}\cmidrule(lr){6-8}\cmidrule(l){9-11}
 & & Mean & Median & P95 & Mean & Median & P95 & Mean & Median & P05 \\
\midrule
\multirow[c]{7}{*}[-3.5pt]{MLP}
 & Single RGB & 8.60 & 8.95 & 12.34 & 42.6 & 42.6 & 68.0 & 27.89 & 27.44 & 23.36 \\
 & Single + learned filter & 7.93 & 8.26 & 10.82 & 36.1 & 36.9 & 59.6 & 29.46 & 28.67 & 24.51 \\
\noalign{\vskip 2pt}
\cdashline{2-11}
\noalign{\vskip 2pt}
 & Reutsky$^{\text{Lee}}$ & 7.57 & 7.89 & 10.51 & 38.3 & 39.3 & 59.6 & 28.99 & 28.11 & 24.52 \\
 & Shrestha$^{\text{Lee}}$ & 7.26 & 7.21 & 10.45 & 37.9 & 39.0 & 55.8 & 29.03 & 28.18 & 25.06 \\
 & Reutsky$^{\text{Omega}}$ & 6.68 & 6.91 & 9.81 & 33.4 & 34.3 & 51.6 & 30.17 & 29.28 & 25.74 \\
 & Shrestha$^{\text{Omega}}$ & 7.87 & 8.18 & 10.88 & 44.8 & 45.2 & 69.6 & 27.70 & 26.95 & 23.16 \\
\rowcolor{DualSpecBlue}
\cellcolor{white}
 & DualSpec (Ours) & \textbf{5.20} & \textbf{4.84} & \textbf{8.20} & \textbf{24.7} & \textbf{27.0} & \textbf{37.8} & \textbf{32.77} & \textbf{31.38} & \textbf{28.54} \\
\midrule
\multirow[c]{7}{*}[-3.5pt]{EDSR}
 & Single RGB & 6.86 & 7.01 & 9.28 & 34.9 & 35.8 & 56.5 & 29.73 & 28.93 & 24.96 \\
 & Single + learned filter & 8.37 & 9.36 & 10.59 & 39.6 & 41.0 & 68.3 & 29.15 & 27.88 & 23.36 \\
\noalign{\vskip 2pt}
\cdashline{2-11}
\noalign{\vskip 2pt}
 & Reutsky$^{\text{Lee}}$ & 5.40 & 5.02 & 8.45 & 31.2 & 31.6 & 58.6 & 31.33 & 30.01 & 24.97 \\
 & Shrestha$^{\text{Lee}}$ & 5.24 & 4.92 & 9.11 & 28.6 & 24.4 & 56.5 & 32.07 & 32.28 & 25.25 \\
 & Reutsky$^{\text{Omega}}$ & 4.72 & 4.39 & 8.60 & 25.6 & 24.3 & 42.9 & 32.84 & 32.31 & 27.44 \\
 & Shrestha$^{\text{Omega}}$ & 7.23 & 5.88 & 16.00 & 37.3 & 31.8 & 80.8 & 30.10 & 29.97 & 22.43 \\
\rowcolor{DualSpecBlue}
\cellcolor{white}
 & DualSpec (Ours) & \textbf{3.64} & \textbf{3.30} & \textbf{5.89} & \textbf{19.7} & \textbf{22.3} & \textbf{29.8} & \textbf{34.92} & \textbf{33.03} & \textbf{30.51} \\
\midrule
\multirow[c]{7}{*}[-3.5pt]{MST++}
 & Single RGB & 6.27 & 6.02 & 8.97 & 33.5 & 28.5 & 61.2 & 30.31 & 30.89 & 24.27 \\
 & Single + learned filter & 7.60 & 7.88 & 9.58 & 40.6 & 42.4 & 67.9 & 28.61 & 27.45 & 23.39 \\
\noalign{\vskip 2pt}
\cdashline{2-11}
\noalign{\vskip 2pt}
 & Reutsky$^{\text{Lee}}$ & 5.22 & 4.87 & 8.42 & 29.2 & 28.1 & 57.5 & 32.00 & 31.04 & 25.27 \\
 & Shrestha$^{\text{Lee}}$ & 5.32 & 5.10 & 9.76 & 29.7 & 21.4 & 68.7 & 32.46 & 33.46 & 23.68 \\
 & Reutsky$^{\text{Omega}}$ & 4.18 & 3.66 & 7.75 & 23.6 & 20.4 & 43.3 & 33.62 & 33.85 & 27.44 \\
 & Shrestha$^{\text{Omega}}$ & 7.18 & 5.13 & 17.75 & 39.1 & 29.8 & 95.6 & 30.26 & 30.54 & 21.35 \\
\rowcolor{DualSpecBlue}
\cellcolor{white}
 & DualSpec (Ours) & \textbf{3.44} & \textbf{3.23} & \textbf{5.32} & \textbf{18.2} & \textbf{20.0} & \textbf{27.6} & \textbf{35.77} & \textbf{33.97} & \textbf{31.17} \\
\bottomrule
\end{tabular}}
\end{table}

% \paragraph{RGB synthesis and evaluation.}
% Each measured spectral cube serves both as ground-truth and as the source from which the corresponding RGB observation is synthesized. For every acquisition configuration, the cube is integrated against that method's sensor--attachment responses and acquired with the noisy acquisition model of the other experiments (capped auto-exposure, Poisson shot noise, dark noise, and 16-bit quantization), independently for each view, and white balanced as the network input.

% Registration is performed independently for each stereo method: SIFT correspondences~\cite{lowe1999object} are extracted from the noise-free RGB renderings of the two views, as for a calibrated rig, and a homography is estimated with RANSAC~\cite{fischler1981random} and applied to the noisy capture of the second view. The left view is warped to the right view, which is used as reference. To enable a fair comparison across baselines, all stereo and single-camera methods are evaluated on the same region within the registered overlap. Table~\ref{tab:supp_real_refL_noisy} repeats the evaluation with the left view as the reference: \ours{} again obtains the best value of every statistic with every backbone.

\paragraph{Evaluation with the left view as reference.}
In the main paper (Table~3), we evaluated spectral reconstruction on displaced-view pairs by registering the left view to the right. Table~\ref{tab:supp_real_refL_noisy} reports the complementary evaluation, in which we register the right view to the left. As in the main experiment, we synthesize noisy RGB observations from the measured spectral cubes and evaluate all methods on a common valid region within the registered overlap. With the left view as reference, \ours{} again achieves the best result for every reported statistic across all three backbones.

\begin{table}[t]
\caption{Displaced-view benchmark without misalignment, left view as the reference: both cameras acquire the reference view, so the stereo input needs no registration. Bold values indicate the best result within each backbone and statistic, while blue shading identifies \ours.}\label{tab:supp_same_view_L}
\centering\scriptsize
\setlength{\tabcolsep}{1.8pt}
\resizebox{\columnwidth}{!}{%
\begin{tabular}{ll*{9}{c}}
\toprule
\multirow[c]{2}{*}[-2.5pt]{SSR} &
\multirow[c]{2}{*}[-2.5pt]{Acquisition / selection} &
\multicolumn{3}{c}{SAM ($^\circ$) $\downarrow$} &
\multicolumn{3}{c}{RMSE $\times 1000\,\downarrow$} &
\multicolumn{3}{c}{PSNR $\uparrow$} \\
\cmidrule(lr){3-5}\cmidrule(lr){6-8}\cmidrule(l){9-11}
 & & Mean & Median & P95 & Mean & Median & P95 & Mean & Median & P05 \\
\midrule
\multirow[c]{7}{*}[-3.5pt]{MLP}
 & Single RGB & 8.60 & 8.95 & 12.34 & 42.6 & 42.6 & 68.0 & 27.89 & 27.44 & 23.36 \\
 & Single + learned filter & 7.93 & 8.26 & 10.82 & 36.1 & 36.9 & 59.6 & 29.46 & 28.67 & 24.51 \\
\noalign{\vskip 2pt}
\cdashline{2-11}
\noalign{\vskip 2pt}
 & Reutsky$^{\text{Lee}}$ & 7.30 & 7.65 & 10.28 & 33.2 & 33.7 & 53.7 & 30.10 & 29.47 & 25.40 \\
 & Shrestha$^{\text{Lee}}$ & 6.91 & 7.06 & 9.90 & 33.1 & 35.0 & 52.5 & 30.09 & 29.13 & 25.60 \\
 & Reutsky$^{\text{Omega}}$ & 6.32 & 6.54 & 9.10 & 28.7 & 27.8 & 47.9 & 31.38 & 31.18 & 26.41 \\
 & Shrestha$^{\text{Omega}}$ & 7.22 & 7.60 & 10.13 & 36.0 & 36.1 & 58.3 & 29.38 & 28.86 & 24.69 \\
\rowcolor{DualSpecBlue}
\cellcolor{white}
 & DualSpec (Ours) & \textbf{4.97} & \textbf{4.73} & \textbf{7.73} & \textbf{21.0} & \textbf{20.2} & \textbf{36.6} & \textbf{34.13} & \textbf{33.89} & \textbf{28.80} \\
\midrule
\multirow[c]{7}{*}[-3.5pt]{EDSR}
 & Single RGB & 6.86 & 7.01 & 9.28 & 34.9 & 35.8 & 56.5 & 29.73 & 28.93 & 24.96 \\
 & Single + learned filter & 8.37 & 9.36 & 10.59 & 39.6 & 41.0 & 68.3 & 29.15 & 27.88 & 23.36 \\
\noalign{\vskip 2pt}
\cdashline{2-11}
\noalign{\vskip 2pt}
 & Reutsky$^{\text{Lee}}$ & 4.37 & 4.23 & 6.81 & 18.1 & 17.2 & 32.1 & 35.40 & 35.28 & 29.89 \\
 & Shrestha$^{\text{Lee}}$ & 3.83 & 3.56 & 6.29 & 17.6 & 16.3 & 30.4 & 35.59 & 35.77 & 30.43 \\
 & Reutsky$^{\text{Omega}}$ & 3.56 & 3.46 & 6.29 & 15.9 & \textbf{13.7} & 28.0 & 36.56 & \textbf{37.25} & 31.19 \\
 & Shrestha$^{\text{Omega}}$ & 4.51 & 4.17 & 7.09 & 20.7 & 19.6 & 36.1 & 34.29 & 34.16 & 28.90 \\
\rowcolor{DualSpecBlue}
\cellcolor{white}
 & DualSpec (Ours) & \textbf{3.01} & \textbf{2.74} & \textbf{4.73} & \textbf{14.5} & \textbf{13.7} & \textbf{25.0} & \textbf{37.38} & 37.24 & \textbf{32.13} \\
\midrule
\multirow[c]{7}{*}[-3.5pt]{MST++}
 & Single RGB & 6.27 & 6.02 & 8.97 & 33.5 & 28.5 & 61.2 & 30.31 & 30.89 & 24.27 \\
 & Single + learned filter & 7.60 & 7.88 & 9.58 & 40.6 & 42.4 & 67.9 & 28.61 & 27.45 & 23.39 \\
\noalign{\vskip 2pt}
\cdashline{2-11}
\noalign{\vskip 2pt}
 & Reutsky$^{\text{Lee}}$ & 4.07 & 3.83 & 6.57 & 17.7 & 17.5 & 30.8 & 35.63 & 35.16 & 30.36 \\
 & Shrestha$^{\text{Lee}}$ & 3.72 & 3.34 & 5.71 & 16.2 & 14.7 & 29.4 & 36.47 & 36.66 & 30.80 \\
 & Reutsky$^{\text{Omega}}$ & 3.27 & 2.93 & 5.18 & 15.1 & 14.1 & 26.8 & 36.90 & 37.04 & 31.59 \\
 & Shrestha$^{\text{Omega}}$ & 4.03 & 3.55 & 6.27 & 18.5 & 16.3 & 33.0 & 35.13 & 35.74 & 29.82 \\
\rowcolor{DualSpecBlue}
\cellcolor{white}
 & DualSpec (Ours) & \textbf{2.85} & \textbf{2.75} & \textbf{4.34} & \textbf{13.3} & \textbf{13.0} & \textbf{21.7} & \textbf{38.17} & \textbf{37.70} & \textbf{33.32} \\
\bottomrule
\end{tabular}}
\end{table}

\begin{table}[t]
\caption{Displaced-view benchmark without misalignment, right view as the reference: both cameras acquire the reference view, so the stereo input needs no registration. Bold values indicate the best result within each backbone and statistic, while blue shading identifies \ours.}\label{tab:supp_same_view_R}
\centering\scriptsize
\setlength{\tabcolsep}{1.8pt}
\resizebox{\columnwidth}{!}{%
\begin{tabular}{ll*{9}{c}}
\toprule
\multirow[c]{2}{*}[-2.5pt]{SSR} &
\multirow[c]{2}{*}[-2.5pt]{Acquisition / selection} &
\multicolumn{3}{c}{SAM ($^\circ$) $\downarrow$} &
\multicolumn{3}{c}{RMSE $\times 1000\,\downarrow$} &
\multicolumn{3}{c}{PSNR $\uparrow$} \\
\cmidrule(lr){3-5}\cmidrule(lr){6-8}\cmidrule(l){9-11}
 & & Mean & Median & P95 & Mean & Median & P95 & Mean & Median & P05 \\
\midrule
\multirow[c]{7}{*}[-3.5pt]{MLP}
 & Single RGB & 8.57 & 8.94 & 12.28 & 42.1 & 42.3 & 67.8 & 27.96 & 27.49 & 23.38 \\
 & Single + learned filter & 7.88 & 8.25 & 10.77 & 35.7 & 35.9 & 59.4 & 29.54 & 28.89 & 24.53 \\
\noalign{\vskip 2pt}
\cdashline{2-11}
\noalign{\vskip 2pt}
 & Reutsky$^{\text{Lee}}$ & 7.26 & 7.65 & 10.23 & 32.7 & 33.0 & 53.6 & 30.18 & 29.63 & 25.42 \\
 & Shrestha$^{\text{Lee}}$ & 6.86 & 7.04 & 9.84 & 32.6 & 34.3 & 52.4 & 30.16 & 29.27 & 25.62 \\
 & Reutsky$^{\text{Omega}}$ & 6.29 & 6.52 & 9.05 & 28.3 & 27.0 & 47.8 & 31.46 & 31.38 & 26.42 \\
 & Shrestha$^{\text{Omega}}$ & 7.17 & 7.58 & 10.08 & 35.5 & 36.1 & 58.1 & 29.47 & 28.85 & 24.73 \\
\rowcolor{DualSpecBlue}
\cellcolor{white}
 & DualSpec (Ours) & \textbf{4.94} & \textbf{4.72} & \textbf{7.68} & \textbf{20.8} & \textbf{19.7} & \textbf{36.7} & \textbf{34.21} & \textbf{34.11} & \textbf{28.78} \\
\midrule
\multirow[c]{7}{*}[-3.5pt]{EDSR}
 & Single RGB & 6.79 & 6.98 & 9.20 & 34.5 & 35.7 & 56.2 & 29.76 & 28.94 & 25.01 \\
 & Single + learned filter & 8.31 & 9.24 & 10.60 & 39.1 & 40.3 & 68.0 & 29.22 & 27.99 & 23.40 \\
\noalign{\vskip 2pt}
\cdashline{2-11}
\noalign{\vskip 2pt}
 & Reutsky$^{\text{Lee}}$ & 4.34 & 4.26 & 6.79 & 17.8 & 16.6 & 32.0 & 35.48 & 35.62 & 29.90 \\
 & Shrestha$^{\text{Lee}}$ & 3.82 & 3.56 & 6.28 & 17.3 & 15.9 & 30.4 & 35.66 & 35.97 & 30.43 \\
 & Reutsky$^{\text{Omega}}$ & 3.56 & 3.45 & 6.34 & 15.7 & 13.3 & 28.1 & 36.64 & 37.53 & 31.16 \\
 & Shrestha$^{\text{Omega}}$ & 4.51 & 4.18 & 7.10 & 20.5 & 19.0 & 36.4 & 34.34 & 34.40 & 28.83 \\
\rowcolor{DualSpecBlue}
\cellcolor{white}
 & DualSpec (Ours) & \textbf{3.00} & \textbf{2.74} & \textbf{4.73} & \textbf{14.4} & \textbf{13.1} & \textbf{25.1} & \textbf{37.44} & \textbf{37.62} & \textbf{32.10} \\
\midrule
\multirow[c]{7}{*}[-3.5pt]{MST++}
 & Single RGB & 6.23 & 6.05 & 8.87 & 33.3 & 28.5 & 60.8 & 30.33 & 30.90 & 24.31 \\
 & Single + learned filter & 7.61 & 7.86 & 9.52 & 40.3 & 41.1 & 67.6 & 28.63 & 27.71 & 23.42 \\
\noalign{\vskip 2pt}
\cdashline{2-11}
\noalign{\vskip 2pt}
 & Reutsky$^{\text{Lee}}$ & 4.03 & 3.86 & 6.51 & 17.4 & 16.7 & 30.8 & 35.75 & 35.54 & 30.35 \\
 & Shrestha$^{\text{Lee}}$ & 3.70 & 3.32 & 5.72 & 16.0 & 13.9 & 29.5 & 36.55 & 37.10 & 30.79 \\
 & Reutsky$^{\text{Omega}}$ & 3.26 & 2.93 & 5.19 & 14.9 & 13.4 & 26.9 & 36.99 & 37.43 & 31.56 \\
 & Shrestha$^{\text{Omega}}$ & 4.02 & 3.55 & 6.24 & 18.3 & 15.9 & 33.2 & 35.19 & 35.95 & 29.76 \\
\rowcolor{DualSpecBlue}
\cellcolor{white}
 & DualSpec (Ours) & \textbf{2.85} & \textbf{2.79} & \textbf{4.33} & \textbf{13.2} & \textbf{12.7} & \textbf{21.7} & \textbf{38.21} & \textbf{37.89} & \textbf{33.32} \\
\bottomrule
\end{tabular}}
\end{table}

\paragraph{Effect of the misalignment.}
To measure how much accuracy the misalignment between the views costs, we repeat the evaluation with both cameras acquiring the reference view, so that the stereo input has no displacement and requires no registration, with the same noise model and evaluation regions. Tables~\ref{tab:supp_same_view_L} and~\ref{tab:supp_same_view_R} report the complete results with the left and right views as the reference. Without misalignment, \ours{} remains the most accurate method with every backbone and both reference views (4.94--4.97$^\circ$, 3.00--3.01$^\circ$, and 2.85$^\circ$ mean SAM with the MLP, EDSR, and MST++), and obtains the best value of every statistic, except for the median PSNR with EDSR and the left reference, where the Reutsky$^{\text{Omega}}$ pair is marginally better (37.25 against 37.24\,dB median PSNR).

Removing the misalignment reduces the mean SAM of every stereo method, except for the Reutsky$^{\text{Lee}}$ pair with the MLP and the right reference, which is unchanged: by 4--6\% with the MLP and 17--19\% with EDSR and MST++ for \ours{}, and by up to 44\% for the selected commercial pairs, the largest reductions being those of the Shrestha pairs with EDSR and MST++ (27--44\%).

\section{Cross-Dataset Evaluation on CAVE}
\label{sec:supp_cave}

The CAVE dataset~\cite{yasuma2010generalized} contains 32 indoor scenes of objects photographed in front of a dark background, which we use as a cross-dataset benchmark (see also Figure~1 of the main manuscript): the attachments and networks trained on Arad-1K are applied without fine-tuning, with the acquisition model and noise of the main experiments. The background is black cloth whose reflectance is close to zero and covers 41\% of the pixels on average. There, the measured spectrum is dominated by the noise of the original acquisition, and the spectral angle, which is undefined for a zero spectrum, becomes unreliable for every method: the background pixels have a mean SAM of 16--41$^\circ$ depending on the configuration, against 6--14$^\circ$ on the objects, and would dominate the image-level metrics. We therefore evaluate on the object pixels only, using segmentation masks of the objects of each scene. Metrics are computed per image over these pixels and then averaged over the 32 scenes, as for the other datasets.

\begin{figure*}[p]
\centering
\includegraphics[width=0.97\textwidth]{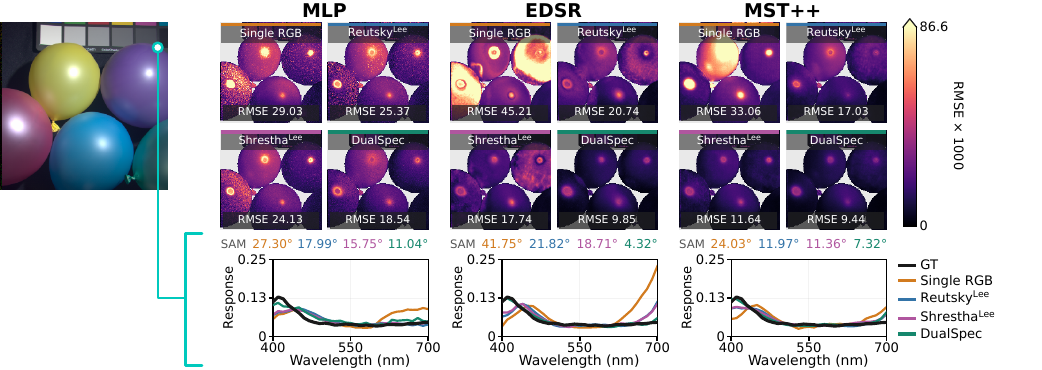}
\includegraphics[width=0.97\textwidth]{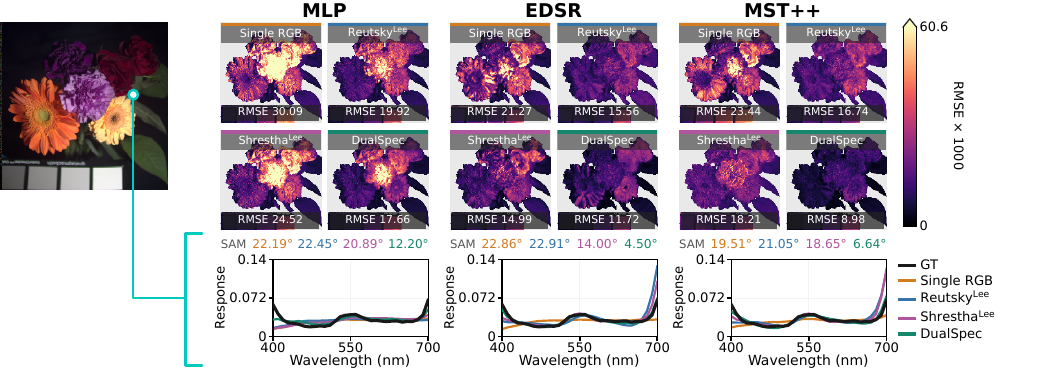}
\includegraphics[width=0.97\textwidth]{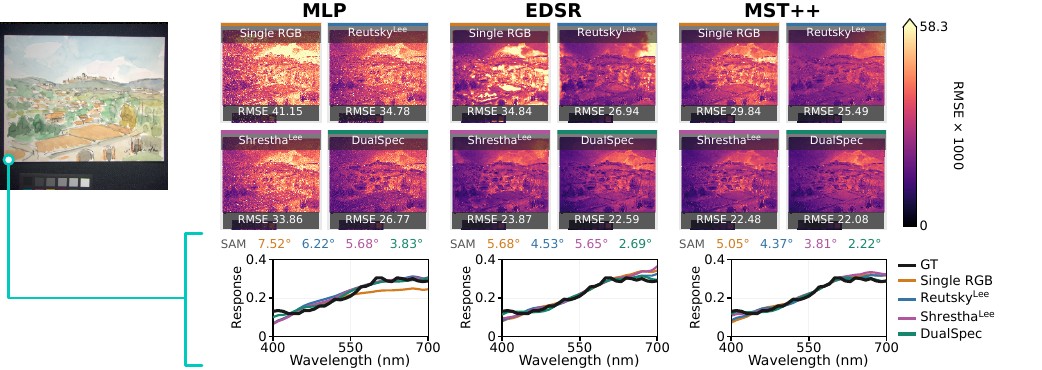}
\caption{Cross-dataset qualitative results on three CAVE scenes, with the Arad-1K models and no fine-tuning. The RMSE maps are restricted to the object pixels of the segmentation masks (background in gray), and the marked pixel is chosen within the objects. Spectra at the marked pixels compare single RGB, two Lee-bank stereo baselines, and \ours{} for all three backbones.}
\label{fig:supp_qualitative_cave}
\end{figure*}

Table~\ref{tab:supp_cave} reports the results. \ours{} achieves the best value of every statistic with every backbone, reducing mean SAM by 32\%, 56\%, and 56\% relative to single RGB for the MLP, EDSR, and MST++, and by 18\%, 29\%, and 26\% relative to the best selected pair. Taken together, the results on CAVE and the displaced-view benchmark show that \ours{} retains its reconstruction advantage on datasets unseen during training.
\Cref{fig:supp_qualitative_cave} shows qualitative results in which the marked pixel is chosen within the objects.

\begin{table}[t]
\caption{Cross-dataset spectral reconstruction on CAVE. Arad-1K models are evaluated without fine-tuning, on the object pixels given by the segmentation masks (the dark background is excluded, see the text). Superscripts identify the filter bank used by each selection method. Bold values indicate the best result within each backbone and statistic, while blue shading identifies \ours.}\label{tab:supp_cave}
\centering\scriptsize
\setlength{\tabcolsep}{1.8pt}
\resizebox{\columnwidth}{!}{%
\begin{tabular}{ll*{9}{c}}
\toprule
\multirow[c]{2}{*}[-2.5pt]{SSR} &
\multirow[c]{2}{*}[-2.5pt]{Acquisition / selection} &
\multicolumn{3}{c}{SAM ($^\circ$) $\downarrow$} &
\multicolumn{3}{c}{RMSE $\times 1000\,\downarrow$} &
\multicolumn{3}{c}{PSNR $\uparrow$} \\
\cmidrule(lr){3-5}\cmidrule(lr){6-8}\cmidrule(l){9-11}
 & & Mean & Median & P95 & Mean & Median & P95 & Mean & Median & P05 \\
\midrule
\multirow[c]{7}{*}[-3.5pt]{MLP}
 & Single RGB & 14.23 & 12.97 & 21.05 & 40.6 & 38.9 & 90.8 & 28.01 & 27.81 & 20.48 \\
 & Single + learned filter & 14.10 & 12.65 & 23.52 & 35.4 & 33.1 & 81.8 & 29.37 & 29.21 & 21.38 \\
\noalign{\vskip 2pt}
\cdashline{2-11}
\noalign{\vskip 2pt}
 & Reutsky$^{\text{Lee}}$ & 12.57 & 11.60 & 18.13 & 31.0 & 30.1 & 55.2 & 30.09 & 29.89 & 24.77 \\
 & Shrestha$^{\text{Lee}}$ & 12.60 & 11.89 & 18.41 & 36.2 & 35.2 & 79.9 & 29.17 & 28.71 & 21.72 \\
 & Reutsky$^{\text{Omega}}$ & 11.77 & 10.87 & 17.11 & 30.6 & 29.3 & 55.2 & 30.30 & 30.07 & 24.79 \\
 & Shrestha$^{\text{Omega}}$ & 12.86 & 12.17 & 18.08 & 39.8 & 37.3 & 86.1 & 28.28 & 28.20 & 21.17 \\
\rowcolor{DualSpecBlue}
\cellcolor{white}
 & DualSpec (Ours) & \textbf{9.71} & \textbf{9.34} & \textbf{14.89} & \textbf{26.3} & \textbf{24.3} & \textbf{52.4} & \textbf{31.74} & \textbf{31.85} & \textbf{25.22} \\
\midrule
\multirow[c]{7}{*}[-3.5pt]{EDSR}
 & Single RGB & 13.99 & 12.91 & 22.11 & 41.6 & 37.7 & 93.1 & 27.88 & 27.94 & 20.28 \\
 & Single + learned filter & 14.42 & 13.72 & 23.41 & 38.2 & 35.2 & 81.2 & 28.63 & 28.43 & 21.45 \\
\noalign{\vskip 2pt}
\cdashline{2-11}
\noalign{\vskip 2pt}
 & Reutsky$^{\text{Lee}}$ & 10.67 & 10.43 & 15.88 & 25.3 & 25.0 & 39.3 & 31.68 & 31.72 & 27.11 \\
 & Shrestha$^{\text{Lee}}$ & 8.68 & 8.26 & 12.73 & 23.4 & 22.3 & 42.4 & 32.50 & 32.54 & 27.06 \\
 & Reutsky$^{\text{Omega}}$ & 9.49 & 9.29 & 14.82 & 24.6 & 23.3 & 45.1 & 31.93 & 31.85 & 26.52 \\
 & Shrestha$^{\text{Omega}}$ & 10.06 & 9.68 & 14.93 & 29.6 & 28.2 & 56.0 & 30.62 & 30.03 & 24.65 \\
\rowcolor{DualSpecBlue}
\cellcolor{white}
 & DualSpec (Ours) & \textbf{6.20} & \textbf{6.23} & \textbf{8.58} & \textbf{18.9} & \textbf{16.8} & \textbf{27.9} & \textbf{34.34} & \textbf{34.48} & \textbf{30.61} \\
\midrule
\multirow[c]{7}{*}[-3.5pt]{MST++}
 & Single RGB & 13.00 & 12.52 & 19.35 & 37.6 & 35.4 & 80.5 & 28.70 & 28.55 & 21.50 \\
 & Single + learned filter & 13.74 & 13.10 & 21.88 & 35.2 & 34.0 & 77.3 & 29.24 & 28.65 & 21.87 \\
\noalign{\vskip 2pt}
\cdashline{2-11}
\noalign{\vskip 2pt}
 & Reutsky$^{\text{Lee}}$ & 11.25 & 10.46 & 17.81 & 25.8 & 26.7 & 43.3 & 31.64 & 30.89 & 26.99 \\
 & Shrestha$^{\text{Lee}}$ & 8.85 & 8.83 & 13.19 & 21.9 & 20.2 & 39.8 & 33.05 & 33.42 & 27.66 \\
 & Reutsky$^{\text{Omega}}$ & 7.67 & 7.45 & 12.48 & 22.4 & 20.1 & 44.6 & 33.17 & 32.95 & 26.62 \\
 & Shrestha$^{\text{Omega}}$ & 9.55 & 10.10 & 12.89 & 26.7 & 24.9 & 51.2 & 31.58 & 31.54 & 25.45 \\
\rowcolor{DualSpecBlue}
\cellcolor{white}
 & DualSpec (Ours) & \textbf{5.66} & \textbf{5.73} & \textbf{8.22} & \textbf{16.9} & \textbf{15.7} & \textbf{26.6} & \textbf{35.40} & \textbf{35.42} & \textbf{30.92} \\
\bottomrule
\end{tabular}}
\end{table}

\section{Additional Qualitative Results}
\label{sec:supp_qualitative}

\begin{figure*}[p]
\centering
\includegraphics[width=0.97\textwidth]{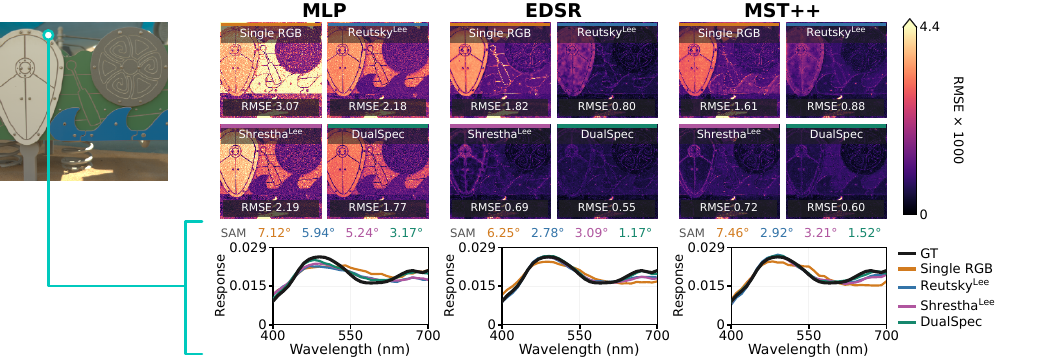}
\includegraphics[width=0.97\textwidth]{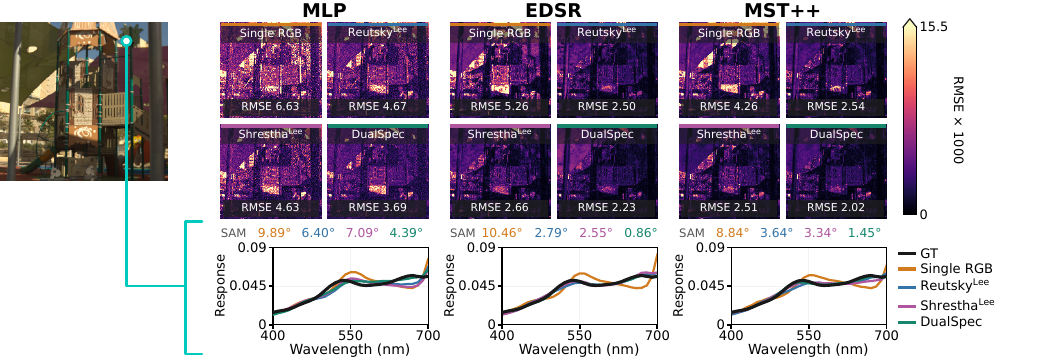}
\includegraphics[width=0.97\textwidth]{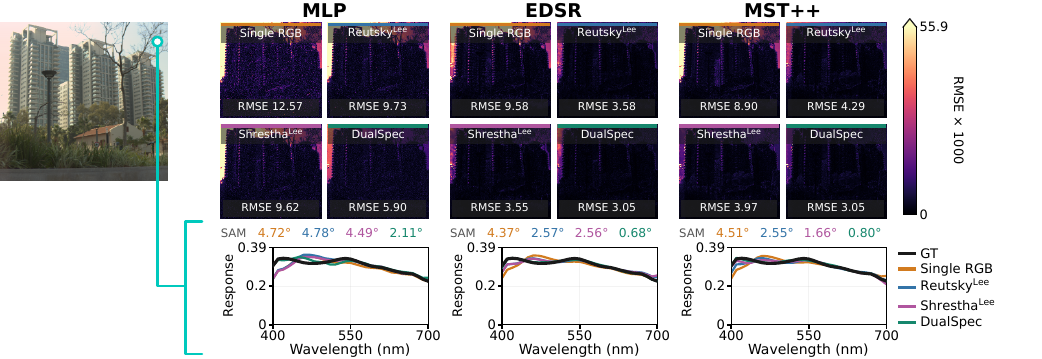}
\caption{Additional qualitative results on three extra Arad-1K test scenes. Spatial RMSE maps and spectra at the marked pixels compare single RGB, two Lee-bank stereo baselines, and \ours{} over 400--700\,nm for all three backbones.}
\label{fig:supp_qualitative_arad}
\end{figure*}

\begin{figure*}[p]
\centering
\includegraphics[width=0.97\textwidth]{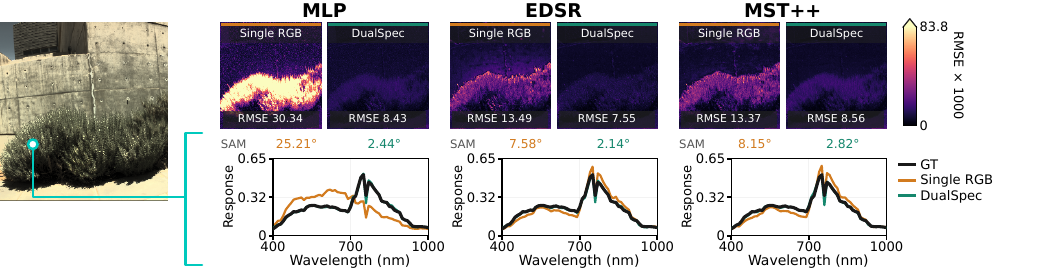}
\includegraphics[width=0.97\textwidth]{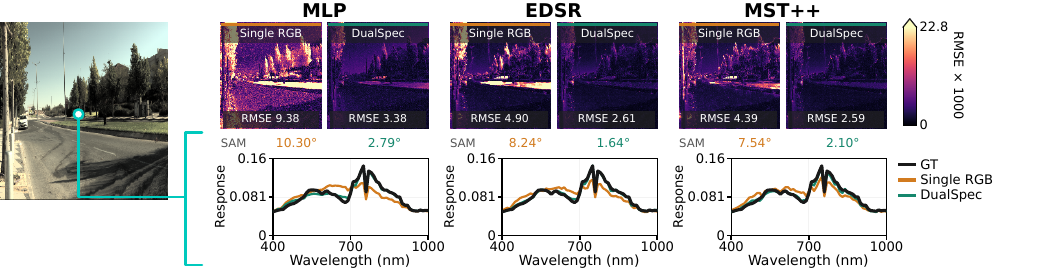}
\includegraphics[width=0.97\textwidth]{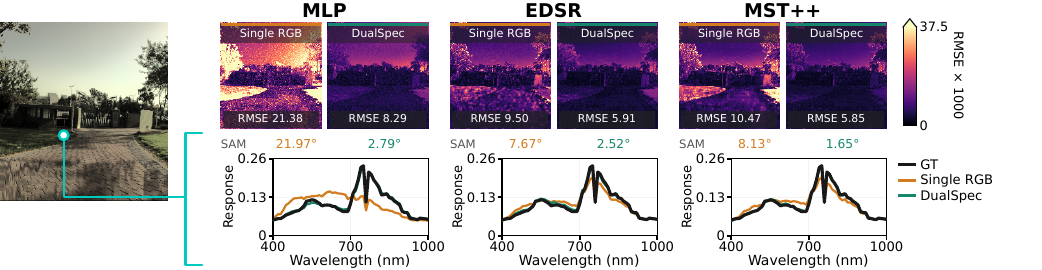}
\caption{Additional qualitative results on three extra BGU-ICVL test scenes. Spatial RMSE maps and spectra at the marked pixels compare single RGB with \ours{} over 400--1000\,nm for all three backbones.}
\label{fig:supp_qualitative_bgu}
\end{figure*}

\begin{figure*}[p]
\centering
\includegraphics[width=0.935\textwidth]{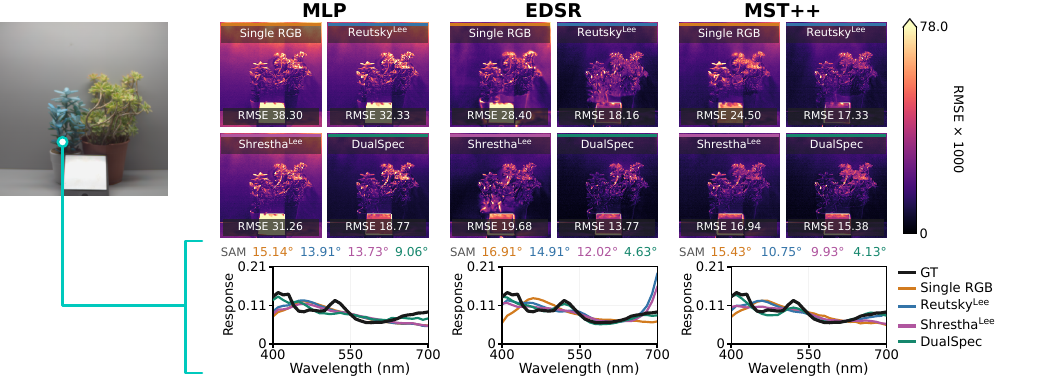}
\includegraphics[width=0.935\textwidth]{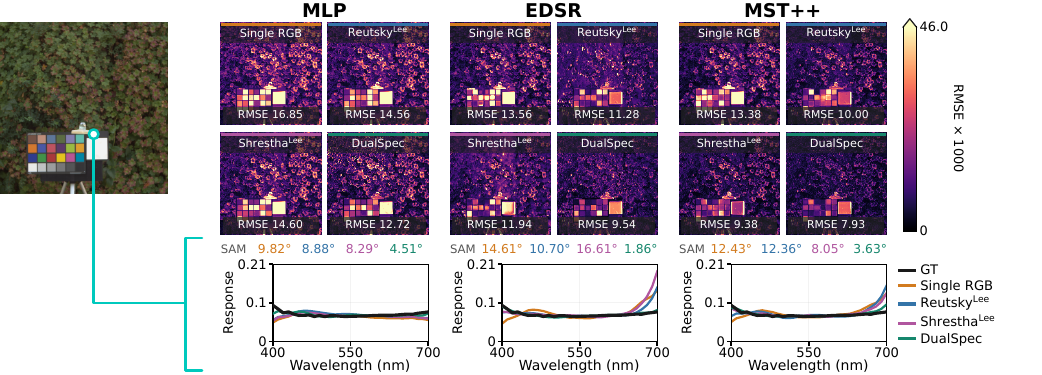}
\includegraphics[width=0.935\textwidth]{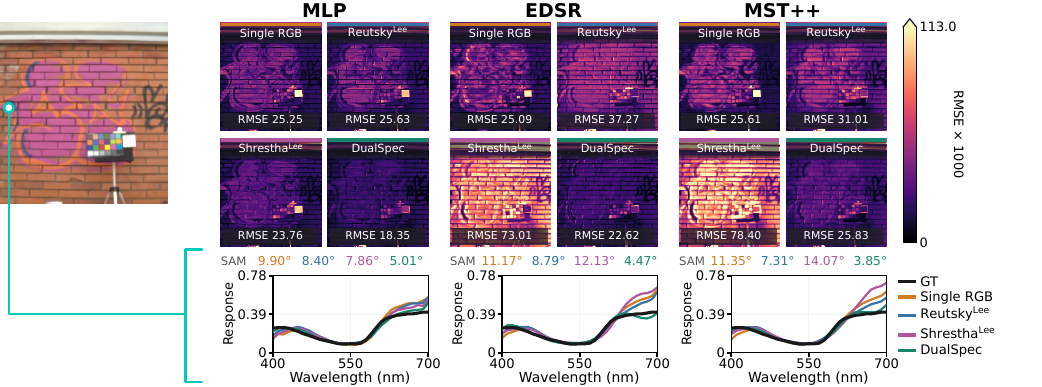}
\caption{Additional qualitative results on three extra scenes of our displaced-view benchmark. Over the registered common region, spatial RMSE maps and spectra at the marked pixels compare single RGB, two Lee-bank stereo baselines, and \ours{} for all three backbones.}
\label{fig:supp_qualitative_real}
\end{figure*}

We show additional qualitative comparisons on Arad-1K, BGU-ICVL, and the displaced-view benchmark. Following the visualization protocol of the main manuscript, each figure shows spatial RMSE maps and reconstructed spectra at a marked pixel for all three backbones. In three visible-range Arad-1K scenes (~\Cref{fig:supp_qualitative_arad}), the complementary measurements reduce reconstruction error and yield spectra closer to the reference, particularly for EDSR and MST++. The three BGU-ICVL examples (~\Cref{fig:supp_qualitative_bgu}) show improvements across the 400--1000\,nm range, including for the pixel-wise MLP, which reconstructs each spectrum without using neighboring pixels. In three independently acquired displaced-view pairs (~\Cref{fig:supp_qualitative_real}), the gains persist after registration in indoor and outdoor scenes with varying depth and texture.

% \section{Additional Qualitative Results}
% \label{sec:supp_qualitative}

% Figures~\ref{fig:supp_qualitative_arad}--\ref{fig:supp_qualitative_real} provide additional qualitative comparisons not included in the main paper. Each figure follows the same visualization protocol as the corresponding main-paper results, showing spatial RMSE maps and reconstructed spectra at a marked pixel for all three reconstruction backbones.

% \paragraph{Arad-1K.}
% Figure~\ref{fig:supp_qualitative_arad} contains three additional visible-range test scenes. Across varied materials and spectral profiles, the complementary measurements reduce spatial error and more closely reproduce the reference spectra, particularly for EDSR and MST++.

% \input{figures/supplementary_figures/supp_qualitative_arad}

% \paragraph{BGU-ICVL.}
% Figure~\ref{fig:supp_qualitative_bgu} shows three additional VNIR examples. The improvements extend across the full 400--1000\,nm interval and remain substantial for the pixel-wise MLP, illustrating the information added by the optimized measurements independently of spatial processing.

% \input{figures/supplementary_figures/supp_qualitative_bgu}

% \paragraph{Displaced spectral views.}
% Figure~\ref{fig:supp_qualitative_real} presents three additional independently acquired scene pairs. The results show that the gains from the learned measurements persist after registration for both indoor and outdoor content with different depth and texture distributions.

% \input{figures/supplementary_figures/supp_qualitative_real}   
{
    \small
    \bibliographystyle{ieeenat_fullname}
    \bibliography{main}
}


\end{document}